\documentclass[preprint,authoryear,11pt]{elsarticle}
\usepackage{amsmath,amsfonts,amssymb, mathtools, blindtext}
\usepackage{url}
\usepackage{booktabs}
\usepackage{multirow}
\usepackage{graphicx}
\usepackage{rotating}
\usepackage{makecell} 
\usepackage{caption}
\usepackage{subcaption}
\usepackage{tabularx}
\usepackage{amssymb}
\usepackage{float}
\usepackage[table,xcdraw]{xcolor}
\usepackage{multirow}
\usepackage{geometry}
\usepackage{threeparttable}
\usepackage[utf8]{inputenc}
\usepackage{enumitem}
\usepackage[T1]{fontenc}
\usepackage{dsfont}
\usepackage{multirow} 

\usepackage{adjustbox} 
\usepackage{hyperref}
\hypersetup{
    colorlinks=true,
    linkcolor=green,
    filecolor=magenta,      
    urlcolor=cyan,
}

\makeatletter
\def\ps@pprintTitle{%
 \let\@oddhead\@empty
 \let\@evenhead\@empty
 \let\@oddfoot\@empty
 \let\@evenfoot\@empty}
\makeatother

\usepackage[ruled,vlined,linesnumbered]{algorithm2e}

\title{Efficient and Privacy-Preserved Link Prediction via Condensed Graphs}

\author[inst1]{Yunbo Long\corref{cor1}}
\ead{yl892@cam.ac.uk}
\author[inst1]{Liming Xu}
\ead{lx249@cam.ac.uk}
\author[inst1,inst2]{Alexandra Brintrup}
\ead{ab702@cam.ac.uk}

\cortext[cor1]{Corresponding author: yl892@cam.ac.uk}

\affiliation[inst1]{
organization={Department of Engineering, University of Cambridge}, 
city={Cambridge},
country={United Kingdom}
}
\affiliation[inst2]{
organization={The Alan Turing Institute}, 
city={London},
country={United Kingdom}
}

\journal{Expert Systems with Applications}

\begin{document}

\begin{abstract}

Link prediction is crucial for uncovering hidden connections within complex networks, enabling applications such as identifying potential customers and products. 
However, this research faces significant challenges, including concerns about data privacy, as well as high computational and storage costs, especially when dealing with large-scale networks.
Condensed graphs, which are much smaller than the original graphs while retaining essential information, has become an effective solution to both maintain data utility and preserve privacy.
Existing methods, however, initialize synthetic graphs through random node selection without considering node connectivity, and are mainly designed for node classification tasks. 
As a result, their potential for privacy-preserving link prediction remains largely unexplored.
We introduce HyDRO\textsuperscript{+}, a graph condensation method guided by algebraic Jaccard similarity, which leverages local connectivity information to optimize condensed graph structures. 
Extensive experiments on four real-world networks show that our method outperforms state-of-the-art methods and even the original networks in balancing link prediction accuracy and privacy preservation.
Moreover, our method achieves nearly 20× faster training and reduces storage requirements by 452×, as demonstrated on the Computers dataset, compared to link prediction on the original networks.
This work represents the first attempt to leverage condensed graphs for privacy-preserving link prediction information sharing in real-world complex networks. It offers a promising pathway for preserving link prediction information while safeguarding privacy, advancing the use of graph condensation in large-scale networks with privacy concerns.
\end{abstract}


\maketitle

\section{Introduction}
\label{sec:introduction}

Complex networks consist of numerous interconnected nodes, often represented as graphs and characterized by properties such as scale-free degree distributions and small-world phenomena \citep{boccaletti2006complex}.
These networks are widely used model complex interactions in real-world scenarios, such as financial, biological, and supply networks. 
In such networks, predicting hidden or emerging connections between nodes---known as {\it link prediction}---is crucial for understanding and reconstructing network dynamics \citep{martinez2016survey}.
For instance, in supply networks, link prediction helps identify potential suppliers or buyers, facilitating to reconstruct supply networks \citep{brintrup2018predicting}. 
In product dependency networks, where items depend on specific components or raw materials, predicting missing or future links improves inventory management and production efficiency by mitigating bottlenecks and ensuring smoother operations \citep{mungo2024reconstructing}. 
Similarly, in co-purchasing networks, which capture relationships between frequently bought-together products, link prediction enhances recommendation systems, enhancing customer experience and driving sales growth \citep{zhang2024heuristic}.

\begin{figure}[t] 
    \includegraphics[width=1\textwidth]{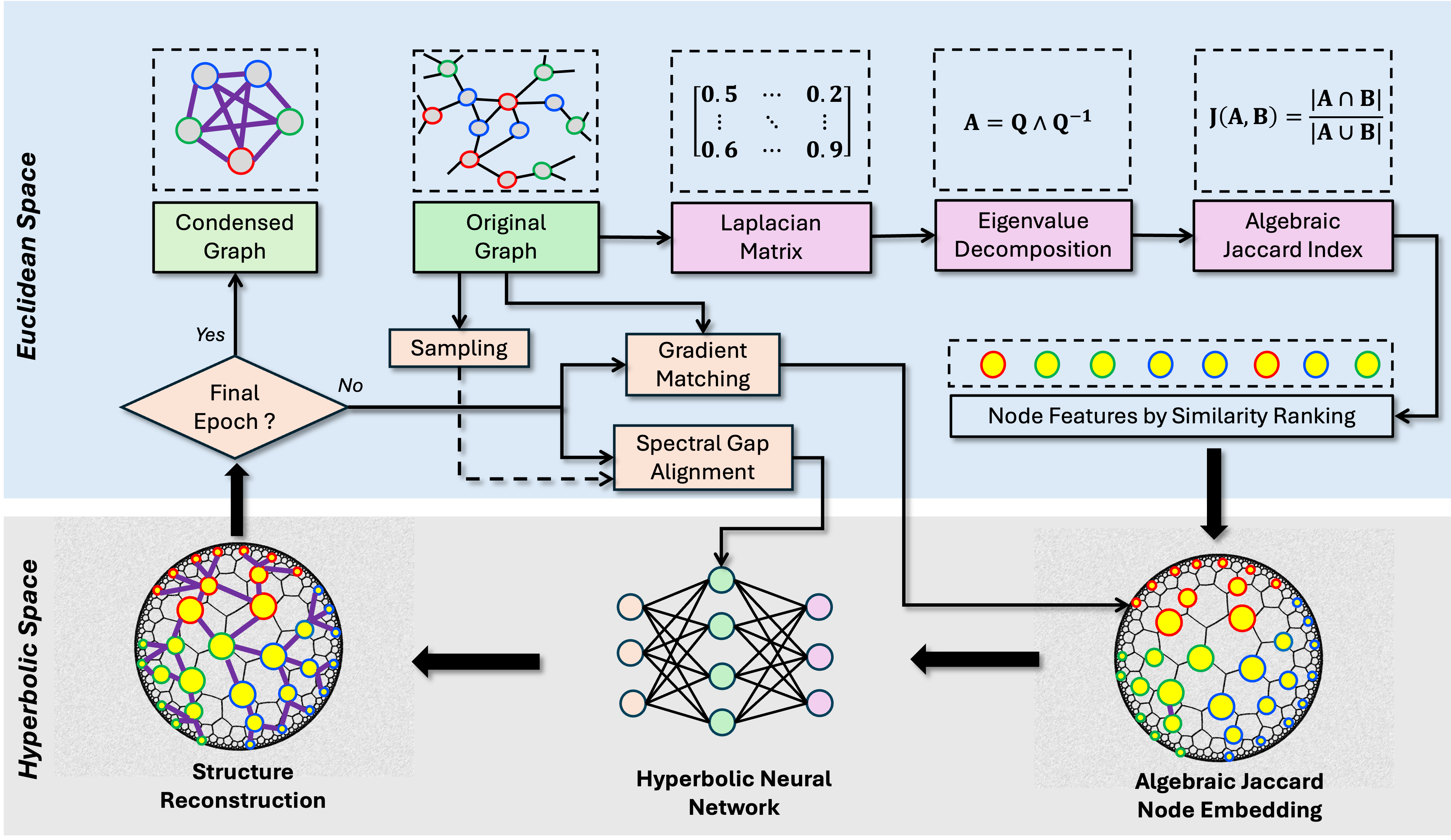}
    \caption{Illustration of the Pipeline of the HyDRO\textsuperscript{+}} 
    \label{fig:workflow} 
\end{figure}

Although link prediction is essential for understanding and analyzing complex networks, its effectiveness often depends on access to comprehensive and high-quality data. 
However, sharing and accessing raw data in real world scenarios is often restricted by commercial confidentiality and privacy regulations \citep{ding2019novel}. 
This is because sensitive information is often implicitly or explicitly embedded in these complex networks, including details about participating entities or customers, as well as edge information such as trades or transactions between companies, which are typically protected by privacy regulations.
Consequently, various privacy-enhancing technologies have been proposed for protecting privacy in graph data, but they still encounter significant limitations.
For instance, although anonymized graphs are widely used, they remain vulnerable to de-anonymization attacks, which can compromise node identities and expose sensitive relationships, ultimately undermining their effectiveness \citep{ji2016graph}.
Moreover, alternative approaches such as graph differential privacy and encryption provide stronger privacy guarantees. 
However, they often introduce excessive noise or computational complexity, reducing the accuracy of link prediction models trained on encrypted graphs and imposing substantial communication and processing overhead \citep{sala2011sharing,chase2010structured}.
Additionally, federated learning, another promising solution, enables decentralized data sharing by allowing multiple clients to collaboratively train models without sharing raw data \citep{rizk2021graph}. However, this approach faces challenges such as data heterogeneity, the complexity of federated feature engineering, and high communication costs during model training \citep{wen2023survey}.
Besides, evaluating whether these PETs can truly preserve graph privacy remains challenging. 
Beyond direct privacy risks related to graph structures, there is growing concern over membership inference attacks (MIAs), where adversaries attempt to infer the presence of specific nodes or links in the original network \citep{wu2022linkteller}. 
Even privacy-enhanced graphs and methods are not immune to such attacks, posing significant risks to sensitive network data.

Another significant challenge in link prediction in complex networks is their large-scale nature, which results in high computational and storage demands \citep{wan2023scalable}. 
The computational cost of link prediction, particularly with Graph Neural Network (GNN)-based architectures, often increases exponentially with network depth \citep{fey2021gnnautoscale}. 
This challenge is even more pronounced in large-scale real-world networks, such as economic networks, which can consist of hundreds of millions of firms connected through billions of supply links \citep{lin2020pagraph, science}. Leveraging GNNs to capture complex dependencies and relational patterns in such networks requires substantial computational resources for both training and inference \citep{abadal2021computing}.
This challenge is further exacerbated by the storage and maintenance requirements of large-scale graph data \citep{ma2025acceleration}. 
For instance, e-commerce recommendation networks, consumer networks, and administrative Value-Added Tax (VAT) networks often consist of billions of nodes and edges, making data storage, sharing, and visualization highly resource-intensive \citep{ma2013large}. Besides, large-scale graph visualization presents another challenge, as effectively representing and interpreting such massive networks is difficult due to their complexity and dense connectivity. Traditional approaches, such as node sampling, often fail to preserve the connectivity and relational dependencies between nodes, thereby limiting their effectiveness in retaining essential information for downstream tasks \citep{knight2012autonetkit}. To overcome these challenges, reducing the complexity of real-world networks in terms of computation, storage, and visualization is crucial.

Recently, graph condensation (GC) has emerged as a promising solution to address these challenges \citep{gong2024gc}. 
It aims to synthesize a {\it compact}, {\it dense} graph to replace the original large-scale, sparse network while maintaining comparable performance on downstream tasks like node classification \citep{hashemi2024comprehensive}. However, for GC methdos, applying graph condensation to link prediction is inherently more challenging than node calssfiatcion, and few studies have explored this direction \citep{gao2024graph}. This is because most existing methods are designed for node classification tasks, which primarily rely on optimizing node features and global structures. In contrast, link prediction depends heavily on local graph structures and connectivity patterns, which are difficult to preserve in condensed graphs.
Currently, only SGDD \citep{yang2024does} and HyDRO \citep{long2025random} have explored the generalization of condensed graphs for link prediction. However, these approaches primarily focus on public networks, such as citation networks, and do not assess the privacy preservation of condensed graphs in real-world network datasets \citep{gao2025graph}. This gap poses a significant risk, as condensed graphs may inadvertently expose sensitive link information, allowing adversaries to infer potential connections between nodes in the original network when applied to link prediction \citep{xu2024survey}.
Furthermore, prior research has demonstrated that the initialization of condensed graphs can accelerate convergence during graph condensation. Nevertheless, state-of-the-art (SOTA) methods, despite their strong generalization capabilities, still rely on random node selection \citep{liu2024gcondenser}. 
This approach is not specifically designed for link prediction and fails to prioritize nodes that play a crucial role in maintaining graph connectivity, thereby limiting its effectiveness in preserving essential structural information.
To address these challenges, it is crucial to develop an enhanced graph condensation method tailored for link prediction, which can effectively balance condensed graph utility and privacy.

To address these gaps, we propose using condensed graphs, which retain essential link prediction information distilled from privacy-sensitive networks through graph condensation techniques, as alternatives to the original graphs for link prediction tasks.
To the best of our knowledge, this is the first such attempt.
Specifically, we begin by reviewing existing graph condensation methods, with an emphasis on their effectiveness in preserving node connectivity for link prediction tasks.
We then introduce HyDRO\textsuperscript{+}, an enhanced version of HyDRO \citep{long2025random}, which replaces its random node selection with a guided approach based on the algebraic Jaccard similarity index.
Extensive experiments on four real-world privacy-sensitive networks demonstrate that HyDRO\textsuperscript{+} outperforms all existing state-of-the-art methods by achieving an optimal balance between link prediction accuracy and privacy preservation, while requiring significantly less computational time and storage space.

The main contributions of this work are summarized as follows:
\begin{itemize}[nosep]
    \item We propose to share link prediction information from privacy-sensitive complex networks through tiny synthetic data generated by graph condensation, enabling secure and efficient network data dissemination for real-world applications.

    \item We present HyDRO\textsuperscript{+}, a graph condensation method based on the \textit{algebraic} Jaccard similarity, which improves upon its predecessor by more effectively preserving structural patterns and local connectivity---key properties essential for link prediction.

    \item 
    Extensive comparison experiments show that the condensed graphs produced by HyDRO\textsuperscript{+} consistently deliver the best or second-best link prediction performance across all four evaluation datasets, achieving at least 95\% accuracy of the original data.
    Notably, these condensed graphs require approximately \textit{one} order of magnitude less computational time and \textit{two} orders of magnitude less storage usage than their original counterparts.
    Moreover, HyDRO\textsuperscript{+} achieves overall best privacy preservation against membership inference attacks, without significantly compromising link prediction accuracy.
    
\end{itemize}

With a good balance between link prediction accuracy and privacy preservation, the condensed graphs generated by HyDRO\textsuperscript{+} would replace their original privacy-sensitive networks for inter-organizational collaboration in link prediction, without concerning privacy leakage.
It thus provides a practical and privacy-aware solution for analyzing and disseminating network data in industrial and commercial settings.

The rest of this paper is structured as follows. 
\autoref{sec:related_work} presents the related work. 
\autoref{sec:approach} details the proposed graph condensation framework for link prediction. The experimental settings and results are presented in \autoref{sec:Settings} and \autoref{sec:results}, respectively. 
\autoref{sec:discussion} discusses the applications and limitations of our method. 
Finally, \autoref{sec:conclusion} concludes the paper and outlines future work.

\section{Related Work} 
\label{sec:related_work}
This section reviews related work, focusing on graph data sharing and privacy preservation, as well as key methodologies for graph condensation.

\subsection{Data Utility and Sharing in Complex Network}
\label{sec:data_sharing }

In the last decade, the rapid development of GNNs and Transformer-based architectures has revolutionized graph learning, enabling significant advancements in solving downstream tasks such as link prediction in complex networks. For instance, Graph Convolutional Networks (GCNs) have proven highly effective in predicting links within social networks by utilizing both node features and graph structure. Building on this foundation, advancements such as Graph Attention Networks (GATs)~\citep{gat} and GraphSAGE~\citep{hamilton2017inductive} have further enhanced the capabilities of graph learning. These methods address critical challenges, including scalability and the handling of heterogeneous graph structures, enabling their application across diverse domains. For instance, they have been successfully employed in recommendation systems~\citep{lakshmi2024link}, biological networks~\citep{ran2024maximum}, and knowledge graphs~\citep{shu2024knowledge}.
Notably, Transformer-based models like SGFormer~\citep{wu2024simplifying} have achieved state-of-the-art performance in predicting connections in citation networks. This model excel at capturing long-range dependencies and hierarchical patterns, demonstrating their potential to push the boundaries of graph-based learning even further.

However, many complex networks—such as VAT networks~\citep{elliott2022supply}, supply networks~\citep{mungo2023reconstructing}, and product networks~\citep{science}—are often commercially sensitive or privately owned, posing significant challenges for data sharing. The intrinsic value and proprietary nature of these datasets make them scarce and highly competitive resources, limiting their availability for research. For instance, economic and supply chain networks frequently contain confidential business information that companies are unwilling to disclose, while product networks may include sensitive customer data or strategic insights that organizations aim to protect.
The sensitive nature of these networks also necessitates rigorous standards for secure data sharing~\citep{science}. Compliance with privacy regulations, such as the General Data Protection Regulation (GDPR), and the potential risks of data breaches further complicate the sharing process. These constraints make it difficult to utilize such networks for research or to standardize them as publicly available datasets~\citep{bier2020methods}. Consequently, researchers often encounter significant barriers in accessing high-quality, real-world network data, which impedes the development, validation, and benchmarking of graph learning methods.
To overcome these challenges, there is an urgent need for privacy-preserving techniques and innovative data-sharing frameworks that facilitate collaborative research while safeguarding sensitive information. Such approaches must strike a balance between the utility of shared data and the protection of proprietary and confidential details. By enabling secure and controlled access to valuable datasets, these methods can empower researchers to advance the field of graph learning without compromising privacy or security.
To address these challenges, there is a critical need for privacy-preserving techniques and innovative data-sharing methods that enable the dissemination of downstream task information—such as link prediction data—derived from original networks, without exposing sensitive or proprietary details. By sharing task-specific insights rather than raw network data, researchers can advance the field of graph learning while maintaining the confidentiality of the underlying networks.

\subsection{Privacy Preservation in Complex Network}
\label{sec: privacy}

Unlike images and tabular data, where anonymization can often be achieved by simply adding noise or perturbations, graph data presents unique challenges due to its inherent structural complexity \citep{zhang2024survey}. 
In graph structures, the relationships between entities (edges) are as crucial as the entities themselves (nodes), and these relationships often carry sensitive information \citep{he2021stealing}. 
For example, in social networks, the connections between individuals can reveal private details about their interactions, while in trading networks, transaction patterns may expose confidential business relationships \citep{dai2024comprehensive}. 
Simply adding noise to node features or edge weights can disrupt the graph’s topology, leading to a substantial loss of structural information, which can cause the graph unsuitable for tasks such as link prediction or community detection \citep{liu2024microservice}. 
Furthermore, even anonymized graphs remain {\it susceptible} to re-identification attacks, in which adversaries exploit structural patterns to infer link identities or reconstruct sensitive connections \citep{wang2023link}. 
These challenges make traditional anonymization techniques inadequate for ensuring privacy in graph data.

Recently, PETs have emerged as promising solutions for secure and efficient data sharing in complex networks. 
These advancements aim to preserve data privacy while enabling collaborative research and analysis. However, each method comes with its own set of challenges, limiting its applicability in practice.
For instance, graph anonymization, which involves removing or obfuscating identifiable information from the graph~\citep{olatunji2024review}, is a straightforward approach but carries a high risk of re-identification. Adversaries can exploit structural patterns to infer sensitive connections or node identities, undermining the privacy guarantees of this method.
To address these limitations, more robust techniques such as graph encryption, differential privacy, and secure multi-party computation have been developed. 
These methods provide stronger privacy guarantees by introducing noise or encrypting data~\citep{hu2024towards}. However, they are often computationally expensive and slow, making them impractical for large-scale or real-time applications.

In addition, federated learning offers a decentralized alternative, allowing multiple parties to collaboratively train models without sharing raw data~\citep{liu2024federated}. While this approach is promising, it requires suitable participants and incurs significant communication costs~\citep{guliani2021training}, which can be prohibitive for large networks.
Given these limitations, there is a growing need to explore innovative methods, such as synthetic data generation~\citep{long2025leveraging}, which can preserve the structural and statistical properties of the original graph while ensuring privacy. Such approaches could provide a more scalable and efficient solution for sharing complex network data without compromising sensitive information.

\subsection{Graph Condensation for Link Prediction}
\label{sec:gc }

Dataset condensation \citep{zhao2020dataset} or distillation \citep{wang2018dataset} has been proposed to generate synthetic data, such as images, while preserving privacy, particularly in sensitive domains like medical dataset sharing \citep{li2022dataset}. Similarly, graph condensation methods, such as GCond, have been developed to condense large graphs with millions of nodes into significantly smaller graphs—sometimes as few as tens of nodes—while achieving comparable performance in node classification tasks to those using the full dataset \citep{jin2021graph}. Building on this, \citep{yang2024does} improved graph condensation by incorporating optimal transportation to learn fundamental spectral properties, such as the Laplacian energy distribution of the condensed graphs. These methods generate smaller graphs with fewer nodes while maintaining the same labels but different feature values, preserving task-specific and structural information while enhancing data privacy through condensed representations. Additionally, \citep{liugraph} addressed the issue of GNN-induced changes to the spectrum of synthetic graphs during condensation, and \citep{zhang2024navigating} enhanced trajectory matching methods for graph condensation, improving performance on larger graphs. These advancements provide new perspectives on graph condensation, offering innovative approaches for generating smaller, privacy-preserving graphs.

Several graph condensation methods have been developed with varying focuses and objectives, while almost all of them are not suitable for tasks like link prediction. Structure-free methods, such as SFGC \citep{zheng2024structure} and GEOM \citep{zhang2024navigating}, prioritize compressing node features and learning node representations through techniques like GNN training trajectory matching. While these methods are effective for node classification, they do not explicitly capture edge prediction information, which is critical for link prediction tasks. Similarly, kernel-based methods like GCSNTK \citep{wang2024fast} (Graph Neural Tangent Kernel) and distribution-based methods like GCDM \citep{liu2022graph} (Graph Distribution Matching) focus on improving computational efficiency and synthesis through kernels and graph embeddings. However, they are not directly optimized for link prediction tasks.
On the other hand, structure-based methods, such as GCond \citep{jin2021graph} and DosCond \citep{doscond}, aim to preserve both node features and graph structure, balancing condensation with structural optimization. While these methods are effective for graph compression and preserving key properties, their loss functions are typically designed for node classification and may not explicitly optimize for link prediction. Additionally, methods like MSGC \citep{msgc} (Multiple Sparse Graphs) use sparse graphs to enhance neighborhood capture. 
In addition, GDEM \citep{liu2023graph} employs eigenbasis matching to preserve global spectral properties, which is valuable for tasks like graph clustering and classification. However, this approach can lead to inaccurate preservation of pairwise node relationships, which are essential for link prediction.

In contrast, there are another two structured-based methods focusing on capturing graph local connectivity with spectral properties, which are particularly well-suited for link prediction tasks. 
For instance, SGDD \citep{yang2024does} leverages Laplacian energy—a measure of overall connectivity—to preserve edge relationships, ensuring that the condensed graph maintains the original graph's edge connectivity. Meanwhile, HyDRO, a state-of-the-art method for link prediction tasks \citep{long2025random}, focuses on the spectral gap, a key indicator of graph connectivity. By preserving this property, HyDRO ensures that the condensed graph retains the connectivity information necessary for predicting new edges. Additionally, HyDRO operates in hyperbolic space, which more accurately reflects graph connectivity and preserves relationships between nodes, especially in graphs with both dense and sparse regions. These spectral-based methods provide a targeted approach for link prediction tasks by maintaining the structural and connectivity properties essential for accurate edge prediction.
Given these advancements, it is essential to evaluate the effectiveness of those structure-based methods in capturing node connections critical for link prediction in real-world complex networks. Such a comparison provides valuable insights into their ability to model the underlying topological and relational information.
Moreover, data initialization plays a crucial role in graph condensation, significantly enhancing convergence speed and boosting performance in node classification tasks \citep{gong2024gc4nc}. Traditionally, data initialization relies on random selection and is often combined with coreset selection and graph coarsening techniques, such as K-Center and Averaging \citep{sun2024gc}. Although these approaches are beneficial for enabling graph condensation models to effectively capture node classification information, they are limited in their ability to preserve the structural properties required for accurate link prediction. Consequently, there is a need for more sophisticated initialization strategies that can better retain relational patterns and topological features, thereby improving the overall performance of link prediction tasks.

\section{HyDRO\textsuperscript{+}: Algebraic Jaccard-Based Graph Condensation}
\label{sec:approach}

\begin{algorithm}[t!]
\SetAlgoVlined
\small
\caption{HyDRO\textsuperscript{+} graph condensation procedure.} 
\label{algo:hydro_plus} 
\textbf{Input:}\( G = ({\bf A}, {\bf X}, {\bf Y}) \) is the original graph; 
$n$ the time of repeated executions; 
$T$ the total number of epochs; 
$l$ learning rate;
$\kappa$ curvature rate;
$\beta$ regularization coefficient;
$C$ a set of classes; and 
$\tau_1$ and $\tau_2$ are number of epochs for optimizing features and structures, respectively. \\
Initialize condensed graph labels $\mathbf{Y'}$ following the original distribution. \\
Compute the Laplacian matrix $\mathbf{L}$ from the adjacency matrix $\mathbf{A}$. \\
Perform eigenvalue decomposition on $\mathbf{L}$ to obtain eigenvectors $\mathbf{V}$. \\
Calculate the algebraic Jaccard similarity $\mathbf{S}_{ij}$ between nodes $i$ and $j$. \\
Compute the average similarity $\bar{\mathbf{S}}$ for each node. \\
Select the top $k$ nodes $\mathbf{X'}$ with the highest average similarity for each label. \\
Initialize $f_\mathrm{hyp}$ for structure learning in hyperbolic space. \\
\For{$i=0, \ldots , n-1$}{
  Randomly initialize a simplified graph convolution network, $\mathrm{SGC}_{\theta}$ \\
  \For{$t=0, \ldots , T - 1$}{
  $\mathcal{L} = 0.0$  \quad\quad\quad\quad\quad \quad\quad\quad\quad\quad
  \quad\quad\quad\quad\quad${\rhd}${total loss} \\
  \For{$c=0, \ldots ,C-1$}{
  $\mathbf{A'} = \mathrm{Sigmoid}\Bigl(\frac{1}{2}\cdot\bigl[{f_\mathrm{hyp}([x^{'}_{i}; x^{'}_{j}]; \Phi; \kappa) + f_\mathrm{hyp}([x^{'}_{j}; x^{'}_{i}]; \Phi; \kappa)}\bigr]\Bigr)$  \\ 
  Sample $({\bf A}_c, {\bf X}_c, {\bf Y}_c)\sim G$ and $({\bf A}_{c}', {\bf X}_{c}', {\bf Y}_{c}')\sim G'$ \\
  Compute $\mathcal{L}_\mathrm{gradient}$  
  \quad\quad\quad\quad\quad \quad\quad\quad\quad\quad${\rhd}${see Equations~\ref{equ:gradient}} 
  \\
  Compute $\mathcal{L}_\mathrm{spectral}$  
  \quad\quad\quad\quad\quad \quad\quad\quad\quad\quad${\rhd}${see Equation~\ref{equ:spectral}}  
  \\
  $\mathcal{L} \leftarrow{} \mathcal{L}_\mathrm{gradient} +  \mathcal{L}_\mathrm{spectral} + \beta\cdot||\mathbf{A'}||_2$
} 
 \If{$\Bigl(k \mod (\tau_1 + \tau_2)\Bigr) < \tau_1$}{
Update ${\bf X'} \leftarrow {\bf X'} -{l_\mathrm{feat}}\cdot 
 \nabla_{\bf X'} \mathcal{L}$ \\
  }
 \Else{
    ${\Phi} \leftarrow {\Phi} -{l_\mathrm{spectral}}\cdot\nabla_{\Phi} \mathcal{L}$\\
 }
$\boldsymbol{\theta}^{G'}_{t+1}\leftarrow \boldsymbol{\theta}^{G'}_{t}-l_\mathrm{SGC}\cdot\nabla_{\theta}\mathcal{L}_{\mathrm{SGC}_{\theta}}$ }}
Update $\mathbf{X'}$ and $\mathbf{A'}$ after training.\\
\textbf{Return:} $({\bf A'}, {\bf X'}, {\bf Y'})$
\end{algorithm}

This section details the pipeline of the proposed HyDRO\textsuperscript{+} method, as illustrated in \autoref{fig:workflow}. 
Based on the HyDRO framework \citep{long2025random}, 
we propose to use Algebraic Jaccard method to guide the selection of representative nodes that best capture the structural properties of the original graph.
To generalize the Jaccard similarity to capture higher-order structural patterns in graphs, algebraic methods are employed. These methods use matrix representations of graphs, such as the adjacency matrix $A$ and the Laplacian matrix $L$, to encode the graph's connectivity and spectral properties.
The algebraic Jaccard similarity is used to quantify the overlap in the neighborhoods of two nodes, which is essential for understanding local structural patterns.
And those selected nodes are embedding to the hyperbolic space for graph structure generation.
This ensures that the condensed graph retains the essential relational and topological features, making it suitable for downstream tasks like link prediction. 
The detailed graph condensation procedure of the HyDRO\textsuperscript{+} is presented in \autoref{algo:hydro_plus}.
We detail this approach in the following sections.

\subsection{Nodes Selection}

Given an original graph \( G = (V, E) \), where \( V \) is the set of nodes and \( E \) is the set of edges, the adjacency matrix of \( G \) is denoted as \( A \in \mathbb{R}^{n \times n} \), where \( n = |V| \). During the condensation process of HyDRO, the sampled nodes and their features are embedded into hyperbolic space to optimize the synthetic graph's features and adjacency. To ensure that the selected nodes preserve the structural and spectral properties of the original graph, we propose a new sampling technique called Algebraic Jaccard-Based Sampling. First, we compute the degree matrix \( D \), which is a diagonal matrix where each entry \( D_{ii} \) represents the degree of node \( i \):
\begin{equation}
D_{ii} = \sum_{j=1}^n A_{ij}, \quad D_{ij} = 0 \text{ for } i \neq j.
\end{equation}
Using the degree matrix \( D \) and the adjacency matrix \( A \), we compute the unnormalized Laplacian matrix \( L \) as:
\begin{equation}
L = D - A.
\end{equation} 
As Laplacian matrix \( L \) is symmetric and positive semi-definite, and its eigenvalue decomposition provides insights into the graph's structure.
To capture the graph's structural patterns, we perform eigenvalue decomposition on \( L \), yielding the eigenvalues \( \lambda_1 \leq \lambda_2 \leq \dots \leq \lambda_n \) and their corresponding eigenvectors \( \mathbf{v}_1, \mathbf{v}_2, \dots, \mathbf{v}_n \). The smallest \( k \) eigenvectors correspond to the most significant structural patterns in the graph, and we select these eigenvectors to form the matrix:
\begin{equation}
\mathbf{V} = [\mathbf{v}_1, \mathbf{v}_2, \dots, \mathbf{v}_k] \in \mathbb{R}^{n \times k}.
\end{equation}
These eigenvectors provide a low-dimensional embedding of the graph, preserving its structural properties. Next, we compute the algebraic Jaccard Similarity, which measures the similarity between nodes based on their embeddings in the eigenvector space. 
The similarity matrix \( S \in \mathbb{R}^{n \times n} \) is computed as:
\begin{equation}
S_{ij} = \frac{\mathbf{v}_i \cdot \mathbf{v}_j}{\|\mathbf{v}_i\| \|\mathbf{v}_j\| + \epsilon},
\end{equation}
where \( \mathbf{v}_i \) and \( \mathbf{v}_j \) are the \( i \)-th and \( j \)-th rows of \( \mathbf{V} \), \( \|\mathbf{v}_i\| \) and \( \|\mathbf{v}_j\| \) are their Euclidean norms, and \( \epsilon \) is a small constant (e.g., \( 10^{-10} \)) to avoid division by zero. To identify structurally central or representative nodes, we compute the average similarity for each node \( i \) as:
\begin{equation}
\bar{S}_i = \frac{1}{n} \sum_{j=1}^n S_{ij}.
\end{equation}
Nodes with high average similarity are considered structurally significant and are prioritized during sampling. The top \( k \) nodes with the highest average similarity are selected to form the sampled subgraph, and the set of selected nodes \( \mathcal{S} \) is given by:
\begin{equation}
\mathcal{S} = \arg\max_{\mathcal{S} \subseteq V, |\mathcal{S}| = k} \sum_{i \in \mathcal{S}} \bar{S}_i,
\end{equation}
where \( \bar{S}_i \) is the average Jaccard similarity for node \( i \). The initialized node features \( \mathbf{X}_\mathcal{S} \) are derived from the selected nodes \( \mathcal{S} \), ensuring that the condensed graph preserves the structural and spectral properties of the original graph.

\subsection{Accelerating Optimization in Hyperbolic Space}

Building on the HyDRO \citep{long2025random}, the initialized node features \( \mathbf{X}_\mathcal{S} \) are embedded into hyperbolic space using the exponential map:
\begin{equation}
\mathbf{X}_\mathcal{S}^{\text{h}} = \mathcal{M}_{h}(\mathbf{X}_\mathcal{S}) \in \mathbb{H}^d,
\end{equation}
where \( \mathcal{M}_{h} \) is the exponential map at the origin \( \mathbf{o} \), as defined previously. 
Specifically, for the node features \( \mathbf{X}_\mathcal{S} \), the exponential map is computed as:
\begin{equation}
\mathcal{M}_{h}(\mathbf{X}_\mathcal{S}) = \tanh(\|\mathbf{X}_\mathcal{S}\|) \frac{\mathbf{X}_\mathcal{S}}{\|\mathbf{X}_\mathcal{S}\|}.
\end{equation}
This initialization accelerates optimization by providing a starting point that is already aligned with the local connectivity patterns of the original graph. The Riemannian gradient descent process is further enhanced by the momentum \( \mu \) and weight decay \( \lambda \), which are adapted to the hyperbolic geometry, ensuring faster convergence and better preservation of structural properties.
Next, edge embeddings are computed by concatenating the hyperbolic feature vectors of node pairs \( (\mathbf{x}'_i,  \mathbf{x}'_j) \), where \( \mathbf{x}'_i, \mathbf{x}'_j \in \mathbf{X}_\mathcal{S}^{\text{h}} \):
\begin{equation}
\mathbf{e}_{ij} = \mathbf{x}'_i \oplus \mathbf{x}'_j \in \mathbb{R}^{2},
\end{equation}
where \( d \) is the dimension of the node features. These edge embeddings capture the structural relationships between nodes in hyperbolic space.
The edge embeddings \( \mathbf{e}_{ij} \) are passed through a hyperbolic neural network \( f_{\text{hyp}} \). This network first applies a Möbius linear transformation \( \mathcal{T}_{\text{Mobius}} \) to the embeddings, then performs hyperbolic batch normalization, and finally uses a hyperbolic ReLU activation. 
The output of this process is a scalar value $a'_{ij}$ representing the predicted edge weight between nodes \( i \) and \( j \):
\begin{equation}
a'_{ij} = f_{\text{hyp}}(\mathbf{e}_{ij};\Phi;\kappa),
\end{equation}
where $\Phi$ is the estimated parameters for the hyperbolic nerual network and $\kappa$ is the curvature rate of the Poincaré ball.
By leveraging \( \mathbf{X}_\mathcal{S}^{\text{h}} \), the edge embeddings \( \mathbf{e}_{ij} \) are  aligned with the local connectivity of the original graph, which accelerates optimization and improves the preservation of structural properties.
The predicted edge weights are then used to construct the symmetric adjacency matrix \( \mathbf{A'} \), where each entry \( \mathbf{A'}_{ij} \) represents the connection probability between nodes \( i \) and \( j \). Then, the adjacency matrix \( \mathbf{A'}_{ij} \) is processed to remove self-loops.
Finally, the adjacency matrix \( \mathbf{A'} \) is normalized, making it suitable for downstream tasks such as link prediction or graph condensation. 
For graph structure optimization, the spectral gaps for the sampled data from the original graphs and synthetic adjacency matrices are defined as $S_{\text{sampling}} = 1 - \lambda_{2,\text{sampling}}$ and $S^{'} = 1 - \lambda^{'}_{2}$, where $\lambda_{2,\text{sampling}}$ and $\lambda^{'}_{2}$ are the second-largest eigenvalues of $A_{\text{sampling}}$ and $A'$, respectively.
To optimize this, we define the loss as below:
\begin{equation}
\label{equ:spectral}
\mathcal{L}_{\text{spectral}} = \left| S^{'} - S_{\text{sampling}} \right|
\end{equation}

\subsection{Gradient Matching}

HyDRO\textsuperscript{+} leverages gradient matching to ensure that models trained on synthetic data \( G' \) and real data \( G \) converge to similar parameter solutions \citep{yang2024does, jin2021graph}. The synthetic data \( G' = \{\mathbf{A'}, \mathbf{X'}, \mathbf{Y'}\} \) is derived from the initialized selected nodes \( \mathcal{S} \), which are transformed into \( G' \) by optimizing the graph structure \( \mathbf{A'} \) and node features \( \mathbf{X'} \) while fixing the node labels \( \mathbf{Y'} \) to match the class distribution of \( \mathbf{Y} \).
And We utilize Simplified Graph Convolution Networks (SGC)~\citep{wu2019simplifyingsgc} for efficient gradient matching.
The objective is to minimize the distance between gradient parameter updates of the SGC on dataset \( G' \) and \( G \):
\begin{equation}
\label{equ:gradient}
    \min_{X',\Phi} \mathbb{E}_{\boldsymbol{\theta_{AJ}} \sim P_{\boldsymbol{\theta_{AJ}}}} \sum_{t=0}^{T-1} \mathrm{dist}_{\cos}\Bigg(
        \nabla_{\boldsymbol{\theta}} \mathcal{L}\Big(\text{SGC}(\boldsymbol{\theta_{t}}, \mathbf{A}, \mathbf{X}), \mathbf{Y}\Big),
        \nabla_{\boldsymbol{\theta}} \mathcal{L}\Big(\text{SGC}(\boldsymbol{\theta_{t}}, \mathbf{f_{hyper}(\Phi)}, \mathbf{X'}), \mathbf{Y'}\Big)
    \Bigg)
\end{equation}
where \( \mathrm{dist}_{\cos}(\cdot, \cdot) \) measures the cosine distance between two vectors (i.e., gradients) to calculate the $\mathcal{L}_\text{gradient}$. And $P_{\theta_{\text{AJ}}}$ is a distribution of initializations based on the Algebraic Jaccard similarity.

\section{Experiments Settings}
\label{sec:Settings}
This section details the experimental settings, including datasets, baseline methods, and evaluation process and metrics.

\subsection{Datasets}
\label{sec:datasets}

\begin{table}[h]
  \centering
  \caption{Statistics of datasets.}
  \resizebox{0.90\linewidth}{!}{
    \begin{tabular}{ccccc|c}
      \toprule
      \textbf{Dataset} & \textbf{Nodes} & \textbf{Edges} & \textbf{Classes} & \textbf{Features} & \textbf{Training/Validation/Test} \\
      \midrule
      Amazon (Computers)       & 13,752   & 491,722    & 10  & 767 & 11,002/1,375/1,375 \\
      Amazon (Photo)   & 7,650   & 238,162    & 8  & 745 & 6,120/765/765 \\
      VAT     & 1,395  & 389,597   & 6  & 100   & 1,117/139/139 \\
      Automotive & 19,389 & 91,709& 7 & 100   & 15,511/1,939/1,939 \\
      \bottomrule
    \end{tabular}
  }
  \label{tab:dataset_details}
\end{table}

We evaluate our methods on four representative real-world complex networks for link prediction while preserving privacy, with detailed network statistics presented in \autoref{tab:dataset_details}. These datasets are particularly relevant because they capture complex relational patterns while requiring strict privacy protection, reflecting real-world challenges in sensitive domains.

\paragraph{Amazon Co-Purchasing Datasets}
The first and second datasets are co-purchasing networks for computers and photography products from Amazon, abbreviated as Computers and Photo \citep{shchur2018pitfalls,zhang2024heuristic}. These datasets capture buying patterns between products, where nodes represent individual products, and edges indicate co-purchasing relationships.

\paragraph{Automotive Dataset}
The third dataset is sourced from Markline \citep{marklines2023}, a global provider of automotive sales data. 
This network captures transactional relationships among companies within the global automotive market. 
In this network, nodes represent entities such as car manufacturers and suppliers, while edges represent transactional relationships, such as a supplier providing components to a car manufacturer. 
This dataset provides a detailed supply relationships within the automotive industry.

\paragraph{VAT Dataset}
The fourth dataset is a product dependency network derived from VAT records. 
In this dataset, nodes represent products, and edges represent dependency relationships, where one product is composed of another. 
For example, a node representing cigars may have a dependency edge to a node representing primary processed tobacco, indicating that cigars are composed of processed tobacco. 
This network captures the hierarchical structure of product assemblies, illustrating how finished goods, such as cigars, are constructed from raw materials and intermediate components.

\vspace{4mm}
These four network datasets capture diverse and complex relationships, including co-purchasing behavior, supply chain transactions, and VAT records, all requiring privacy protection. Their varied structural properties ensure a rigorous evaluation of our method across different network topologies, making the results broadly applicable to real-world scenarios.

\begin{table}[h]
\centering
\caption{Datasets with different reduction rates ($r$) under transductive settings.}
\label{tab:reduction_rate_statistics}
\resizebox{0.80\textwidth}{!}{%
\begin{tabular}{cccc}
\toprule
  {\bf Dataset} &
  {\bf Labeling Rate} &
  \begin{tabular}[c]{@{}c@{}}{\bf Reduction Rate $r$(\%)}\end{tabular} &
  \multicolumn{1}{c}{\begin{tabular}[c]{@{}c@{}}{\bf Reduction Rate (Labeled Nodes)}\end{tabular}} \\ \midrule
\multirow{3}{*}{Computers}  & \multirow{3}{*}{80\%} & 0.4\%  & 0.5\%  \\
                            &                       & 0.8\%  & 1\%    \\
                            &                       & 1.6\%  & 2\%    \\ \hline
\multirow{3}{*}{Photo}      & \multirow{3}{*}{80\%} & 0.8\%  & 1\%    \\
                            &                       & 1.6\%  & 2\%    \\
                            &                       & 4.0\%  & 5\%    \\ \hline
\multirow{3}{*}{VAT}        & \multirow{3}{*}{80\%} & 1.5\%  & 1.9\%  \\
                            &                       & 3.0\%  & 3.8\%  \\
                            &                       & 6.0\%  & 7.5\%  \\ \hline
\multirow{3}{*}{Automotive} & \multirow{3}{*}{80\%} & 0.11\% & 0.14\% \\
                            &                       & 0.56\% & 0.7\%  \\
                            &                       & 1.12\% & 1.4\%  \\ \bottomrule
\end{tabular}%
}
\label{tab:dataset_reduction}
\end{table}

\subsection{Baselines}\label{sec:baselines}

We compare our model against several baselines, categorized into two groups. 
The first group consists of traditional graph reduction methods, including coreset-based approaches such as Random, Herding~\citep{welling2009herding}, and KCenter~\citep{sener2017active}. 
These methods aim to reduce graph size while preserving structural features, making them suitable for link prediction tasks. 
The second category includes structured-based graph condensation methods, which are specifically designed to preserve the structural information of the original graphs while producing graphs with much smaller sizes. 
In this category, we consider 
GCOND~\citep{jin2021graph}, 
MSGC~\citep{msgc}, 
DosCond~\citep{doscond}, 
GDEM~\citep{liugraph}, 
SGDD~\citep{yang2024does}, and 
HyDRO~\citep{long2025random} as baseline methods.
These approaches are particularly relevant as they focus on preserving essential graph properties while enabling effective link prediction.
Among them, GCOND, MSGC, and DosCond primarily emphasize preserving node features and graph structure, while GDEM, SGDD, and HyDRO focus on maintaining spectral properties and connectivity---both of which are crucial for capturing relational patterns in complex networks.

\subsection{Settings and Hyperparameters}

\paragraph{Experimental Settings}
The specific dataset splitting and reduction rates for each dataset are provided in \autoref{tab:dataset_details} and \autoref{tab:dataset_reduction}, respectively. 
We followed the same procedures as described in \citet{jin2021graph} for graph preprocessing. 
The experiments involved the following steps: (1) the graph condensation models were first trained on the \textit{training} datasets. (2) The generated condensed graphs were selected based on their link prediction performance on the validation datasets. (3) These selected condensed graphs were then used to train a GCN model for link prediction, and its performance was evaluated on the test dataset. (4) The same condensed graphs were evaluated for additional downstream tasks, such as membership inference attacks on nodes and links.

\paragraph{Hyperparameter Tuning}
We followed the same setting as in \citet{long2025random} to configure the hyperbolic neural network in HyDRO\textsuperscript{+}. 
For other baseline methods, we adopted the model structures and key parameters as specified in their respective original papers.
To ensure a fair comparison, we performed hyperparameter finetuning for all graph condensation methods.
For all models, we fine-tuned the learning rate for both feature $l_\mathrm{feat}$ and structure  $l_\mathrm{spectral}$ optimization within the set $\{.1, .01, .001, .0001\}$. 
The outer loop was fixed at 10 and the inner loop at 1. 
The number of hidden layers was fixed at 2, while the number of hidden units was set to 256. 
GNN-based methods such as SGC and GCN were equipped with 2 layers with 256 hidden units, and the total training epoch was set to 600. 
The learning rate $l_\mathrm{SGC}$ for these models was set to 0.01.
For HyDRO and HyDRO\textsuperscript{+}, the regularization coefficient $\beta$ was set to 0.1, the parameters $\tau_1$ and $\tau_2$ were fixed at 40 and 10, respectively, and the curvature rate $\kappa$ was set to -0.1.
The hyperbolic neural network was configured with two hidden layers, each containing 256 hidden units.

For each dataset, the hyperparameters were fine-tuned based on the configuration that achieved the highest reduction rate. Once selected, these hyperparameters were applied consistently across all reduction rates for the same dataset to ensure uniformity in the evaluation and reduce the finetuning cost. For methods with special parameters requiring fine-tuning, we adhered to the settings specified in their original papers, maintaining consistency with their established implementations.

\subsection{Evaluation Framework}
\label{sec:evaluation}

\begin{figure}[t] 
    \includegraphics[width=1\textwidth]{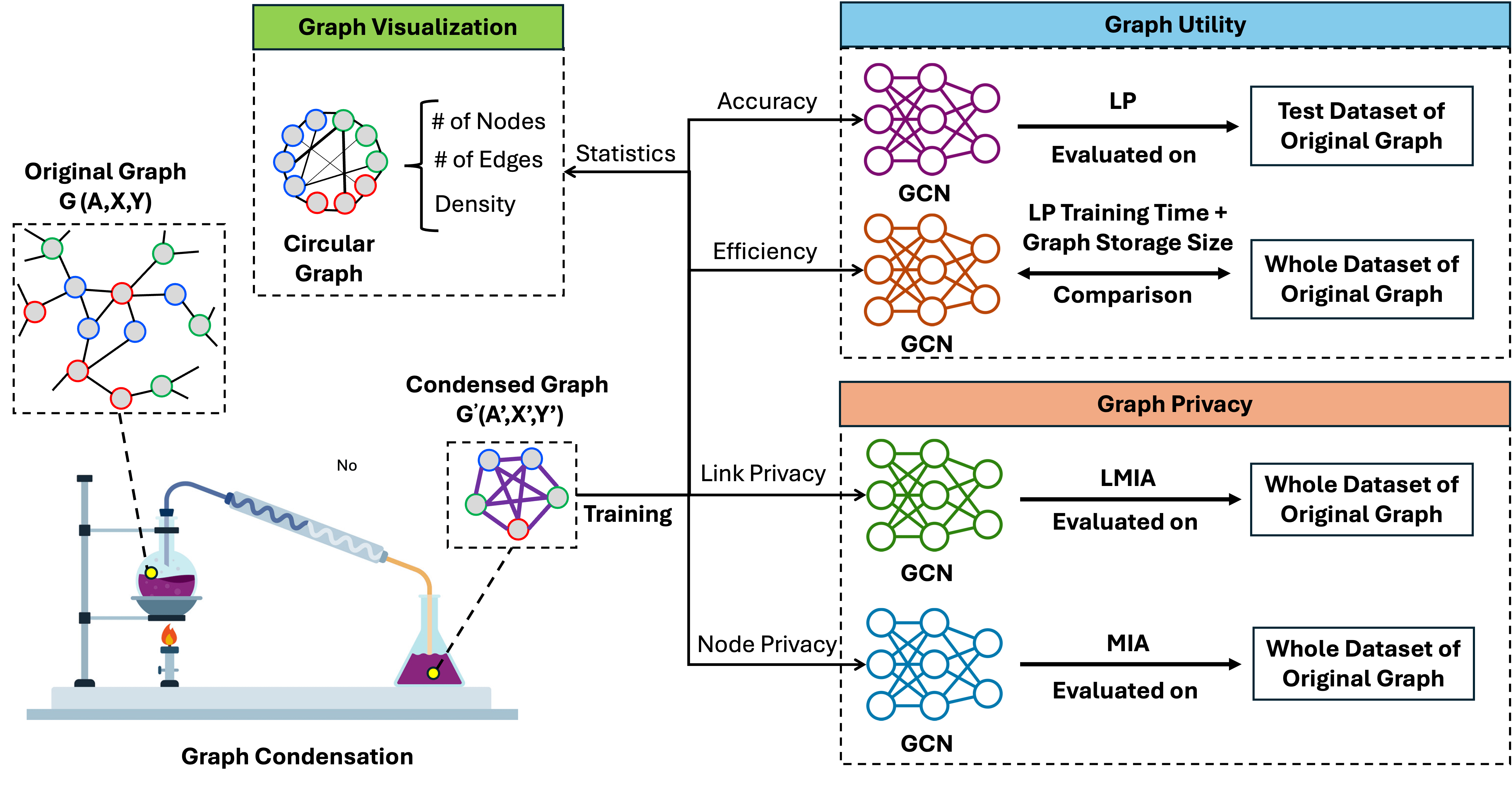}
    \caption{Illustration of the evaluation process of an condensed graph} 
    \label{fig:evaluation} 
\end{figure}

We propose an evaluation framework, as illustrated in \autoref{fig:evaluation}, to assess the condensed graphs from three key aspects: graph privacy, utility, and statistics. This framework is designed to comprehensively evaluate the effectiveness of graph condensation in preserving sensitive information while maintaining the structural and functional properties of the original graph.
For data processing, we adopt the transductive settings  proposed by \citet{jin2021graph}.
For link prediction, we train a graph neural network (GNN) with two hidden layers and 128 hidden units, using a learning rate of 0.001 and 100 epochs. The dataset is split into training, validation, and test sets with a ratio of 0.7, 0.1, and 0.2, respectively. The model is trained on the training set and evaluated on the validation set. For testing, we measure accuracy using the F1 score, which balances precision and recall, reflecting the proportion of correctly predicted edges (both positive and negative). To ensure robustness, the results are averaged over 10 runs, with the mean and standard deviation of the test accuracy reported.

\paragraph{Graph Utility}
To assess the utility of the condensed graphs, we evaluate their performance on general downstream tasks, such as link prediction. 
Specifically, we train link prediction models on both the original and condensed graphs and compare their predictive accuracy. 
In addition, we analyze computational efficiency by measuring the training time and storage usage associated with using condensed graphs versus original graphs. 
This evaluation ensures that condensation methods provide practical benefits while maintaining high performance in graph-based tasks.

\paragraph{Graph Statistics and Visualization}
To understand how well the condensed graphs retain the structural properties of the original data, we perform a direct comparison of graph statistics and structures. 
By visualizing key graph attributes, we examine whether the condensation process disrupts structural fidelity enough to protect the graph structures information.

\paragraph{Graph Privacy}
To assess privacy risks in condensed graphs, we employ Membership Inference Attack (MIA) \citep{duddu2020quantifying} and Link Membership Inference Attack (LMIA) \citep{wang2023link}. 
MIA attempts to determine whether a specific node was included in the training data, while LMIA focuses on inferring whether an edge was part of the original graph. 
Since link prediction models may be vulnerable to structural information leakage, LMIA is particularly important. 
We conduct both attacks over 10 runs with different random seeds to ensure robustness. 
MIA results are reported using attacker success rate (accuracy), while LMIA performance is measured using F1 score. 
The results, presented as mean and standard deviation, provide insights into the trade-off between utility and privacy protection.

\section{Experimental Results}
\label{sec:results}
This section evaluates the performance of condensed graphs generated by our proposed method and baseline approaches in link prediction tasks, graph data privacy preservation and computational and storage efficiency. 
Each evaluation is conducted \textbf{ten times}, with the results reported as the mean and standard deviation in the format $\text{mean} \pm \text{std dev}$.

\subsection{Link Prediction}
\begin{table}[t!]
\renewcommand{\arraystretch}{1.5}
\centering
\caption{
    Experimental results on {\bf link prediction (LP)}. 
    The results are shown in the format of ({\tt mean $\pm$  std dev}).
    Higher values indicate better performance. 
    The best and second-best results are highlighted in \textcolor{blue}{\textbf{blue}} and \textbf{black boldfaced}, respectively.
}
\label{tab:link_prediction}
\resizebox{\textwidth}{!}{%
\begin{tabular}{ccccccccccclc}
\toprule
\multirow{2}{*}{\textbf{Datasets}} &
  \multirow{2}{*}{\textbf{\begin{tabular}[c]{@{}c@{}}$r$ (\%)\end{tabular}}} &
  \multicolumn{3}{c}{\textbf{Traditional Methods}} &
  \multicolumn{7}{c}{\textbf{Structure-based Graph Condensation Methods}} &
  \multirow{2}{*}{\textbf{\begin{tabular}[c]{@{}c@{}}Whole \\ Dataset\end{tabular}}} \\ \cmidrule(lr){3-5} \cmidrule(lr){6-12}
  &
  &
  Random &
  Herding &
  KCenter &
  GCond &
  SGDD &
  DosCond &
  MSGC &
  GDEM &
  HyDRO &
  HyDRO\textsuperscript{$\mathbf{+}$} &
   \\ \midrule
\multirow{3}{*}{Computers} &
  0.40\% &
  52.63±0.95 &
  52.59±0.83 &
  55.10±1.69 &
  50.14±0.07 &
  72.02±0.32 &
  52.24±1.18 &
  58.66±4.27 &
  51.31±1.42 &
  \textcolor{blue}{\textbf{73.01±1.23}} &
  \textbf{72.61±1.12} &
  \multirow{3}{*}{75.59±0.82} \\ \cline{2-12}
 &
  0.80\% &
  54.70±2.18 &
  54.14±1.40 &
  62.76±4.28 &
  50.99±0.51 &
  72.83±0.29 &
  66.95±0.72 &
  54.56±3.31 &
  50.72±0.06 &
  \textcolor{blue}{\textbf{72.93±0.14}} &
  \textbf{72.88±0.23} &
   \\ \cline{2-12}
 &
  1.60\% &
  55.61±1.16 &
  55.05±1.13 &
  65.05±1.99 &
  66.82±0.32 &
  \textbf{72.44±0.56} &
  51.76±1.71 &
  55.13±7.74 &
  50.89±0.51 &
  71.89±0.48 &
  \textcolor{blue}{\textbf{72.76±0.65}} &
   \\ \hline
\multirow{3}{*}{Photo} &
  0.80\% &
  52.63±0.99 &
  53.51±1.25 &
  68.43±3.04 &
  64.44±2.42 &
  \textbf{72.78±1.92} &
  66.61±3.22 &
  66.80±6.34 &
  50.01±0.01 &
  72.66±1.61 &
  \textcolor{blue}{\textbf{73.11±1.43}} &
  \multirow{3}{*}{77.36±0.99} \\ \cline{2-12}
 &
  1.60\% &
  60.85±2.84 &
  57.71±1.29 &
  71.28±0.67 &
  69.31±1.62 &
  \textcolor{blue}{\textbf{74.87±2.12}} &
  58.60±3.23 &
  61.69±5.97 &
  50.28±0.02 &
  \textbf{74.56±0.74} &
  74.30±1.08 &
   \\ \cline{2-12}
 &
  4.00\% &
  62.24±4.69 &
  64.55±1.69 &
  73.58±1.28 &
  67.87±1.87 &
  74.22±2.69 &
  58.60±3.23 &
  67.20±8.09 &
  50.02±0.01 &
  \textcolor{blue}{\textbf{76.10±0.97}} &
  \textbf{75.86±0.87} &
   \\ \hline
\multirow{3}{*}{VAT} &
  1.50\% &
  57.98±1.08 &
  58.62±2.55 &
  60.27±2.29 &
  \textcolor{blue}{\textbf{67.39±0.85}} &
  62.55±1.52 &
  64.84±1.56 &
  60.21±2.13 &
  50.96±0.90 &
  63.63±1.66 &
  \textbf{66.61±0.83} &
  \multirow{3}{*}{65.64±1.21} \\ \cline{2-12}
 &
  3.00\% &
  62.82±1.24 &
  60.21±2.40 &
  63.35±1.85 &
  61.06±1.27 &
  60.48±2.86 &
  \textbf{68.51±0.36} &
  60.69±2.28 &
  53.19±0.50 &
  64.84±2.04 &
  \textcolor{blue}{\textbf{68.71±0.25}} &
   \\ \cline{2-12}
 &
  6.00\% &
  65.90±1.64 &
  66.17±2.24 &
  62.55±1.94 &
  63.56±1.19 &
  63.99±0.78 &
  65.16±0.73 &
  60.48±2.18 &
  53.72±1.55 &
  \textcolor{blue}{\textbf{68.62±0.56}} &
  \textbf{68.16±0.13} &
   \\ \hline
\multirow{3}{*}{Automotive} &
  0.11\% &
  62.05±2.34 &
  61.52±1.51 &
  58.44±2.20 &
  \textcolor{blue}{\textbf{68.13±0.62}} &
  66.57±1.17 &
  54.69±1.83 &
  67.71±0.69 &
  58.88±1.77 &
  66.37±0.61 &
  \textbf{67.88±0.79} &
  \multirow{3}{*}{70.06±0.36} \\ \cline{2-12}
 &
  0.56\% &
  \bf67.90±0.31 &
  64.36±2.29 &
  63.64±2.25 &
  67.26±0.63 &
  66.55±0.11 &
  64.42±1.49 &
  63.86±1.58 &
  61.98±1.81 &
  66.94±0.76 &
  \textcolor{blue}{\textbf{69.13±0.15}} &
   \\ \cline{2-12}
 &
  1.12\% &
  66.27±1.58 &
  63.27±2.58 &
  65.30±1.88 &
  64.25±0.69 &
  65.32±0.36 &
  64.42±0.81 &
  65.59±0.539 &
  63.49±1.32 &
  \textcolor{blue}{\textbf{67.49±0.39}} &
  \textbf{66.50±0.56} &
  \\ \bottomrule
\end{tabular}%
}
\end{table}

\autoref{tab:link_prediction} presents the link prediction performance of using the condensed graphs produced by various graph condensation methods.
The results demonstrate that HyDRO\textsuperscript{+} and HyDRO achieve the best performance in most cases. 
However, HyDRO\textsuperscript{+} outperforms HyDRO by consistently ranking first or second across four datasets at various reduction rates.
In addition, our results show that HyDRO\textsuperscript{+} achieves link prediction performance close to that of the original graphs and even surpasses the performance of the original graphs in the VAT dataset at all reduction rates. 
Specifically, HyDRO\textsuperscript{+} achieves 68.71\% accuracy on the VAT dataset at a 3\% reduction ratio, outperforming the original dataset by 4.7\%. 
This demonstrates the effectiveness of HyDRO\textsuperscript{+} in preserving essential link prediction information despite significant graph compression.

Additionally, SGDD performs well, particularly on the Computers and Photo datasets. 
However, its performance is less impressive on the VAT and Automotive datasets, where it falls short compared to HyDRO\textsuperscript{+} and HyDRO.
In contrast, other graph condensation methods, such as GCond, DosCond, MSGC, and GDEM, fail to deliver consistent performance across all four network datasets. 
These methods, primarily designed to capture the global structure of networks, struggle to effectively distill the link prediction information necessary for accurate predictions. 
In some cases, these methods even underperform compared to random node and edge selection approaches for link prediction tasks.
Overall, the results highlight that HyDRO\textsuperscript{+}, HyDRO and SGDD are effective at distilling relevant nodes connection information into their condensed graphs, making them better choices for tasks involving link prediction in networks.

\begin{table}[t]
\centering
\caption{Comparison of statistics between the original graphs and the condensed graphs generated by HyDRO\textsuperscript{+}.}
\label{tab:statistics}
\resizebox{0.9\textwidth}{!}{%
\begin{tabular}{ccccccccc}
\toprule
 &
  \multicolumn{2}{c}{\textbf{Computers ($r$=0.4\%)}} &
  \multicolumn{2}{c}{\textbf{Photo ($r$=0.8\%)}} &
  \multicolumn{2}{c}{\textbf{VAT ($r$=1.5\%)}} &
  \multicolumn{2}{c}{\textbf{Automotive ($r$=0.11\%)}} \\ 
  \cmidrule(lr){2-3} \cmidrule(lr){4-5} \cmidrule(lr){6-7} \cmidrule(lr){8-9}
\multirow{-2}{*}{} &
  Original &
  \cellcolor[HTML]{EFEFEF}Condensed &
  Original &
  \cellcolor[HTML]{EFEFEF}Condensed &
  Original &
  \cellcolor[HTML]{EFEFEF}Condensed &
  Original &
  \cellcolor[HTML]{EFEFEF}Condensed \\ \midrule
\# of Nodes &
  13,752 &
  \cellcolor[HTML]{EFEFEF}55 &
  7,650 &
  \cellcolor[HTML]{EFEFEF}61 &
  1,395 &
  \cellcolor[HTML]{EFEFEF}21 &
  19,389 &
  \cellcolor[HTML]{EFEFEF}22 \\
\# of Edges &
  491,722 &
  \cellcolor[HTML]{EFEFEF}1,431 &
  238,162 &
  \cellcolor[HTML]{EFEFEF}1,830 &
  389,597 &
  \cellcolor[HTML]{EFEFEF}210 &
  91,709 &
  \cellcolor[HTML]{EFEFEF}231 \\
Density &
  0.52\% &
  \cellcolor[HTML]{EFEFEF}96.36\% &
  0.81\% &
  \cellcolor[HTML]{EFEFEF}100\% &
  40\% &
  \cellcolor[HTML]{EFEFEF}100\% &
  0.05\% &
  \cellcolor[HTML]{EFEFEF}100\% \\ \bottomrule
\end{tabular}%
}
\end{table}

\subsection{Statistics and Visualization}

\begin{figure}[t]
    \centering
    \footnotesize
    \begin{tabular}{@{}c@{}c@{}c@{}c@{}c@{}c@{}}
        & GCond & SGDD & GDEM & HyDRO & HyDRO\textsuperscript{+} \\[0.00ex] 
        \rotatebox{90}{\makecell{Computers\\($r$=0.4\%)}} & 
        \includegraphics[width=0.18\textwidth]{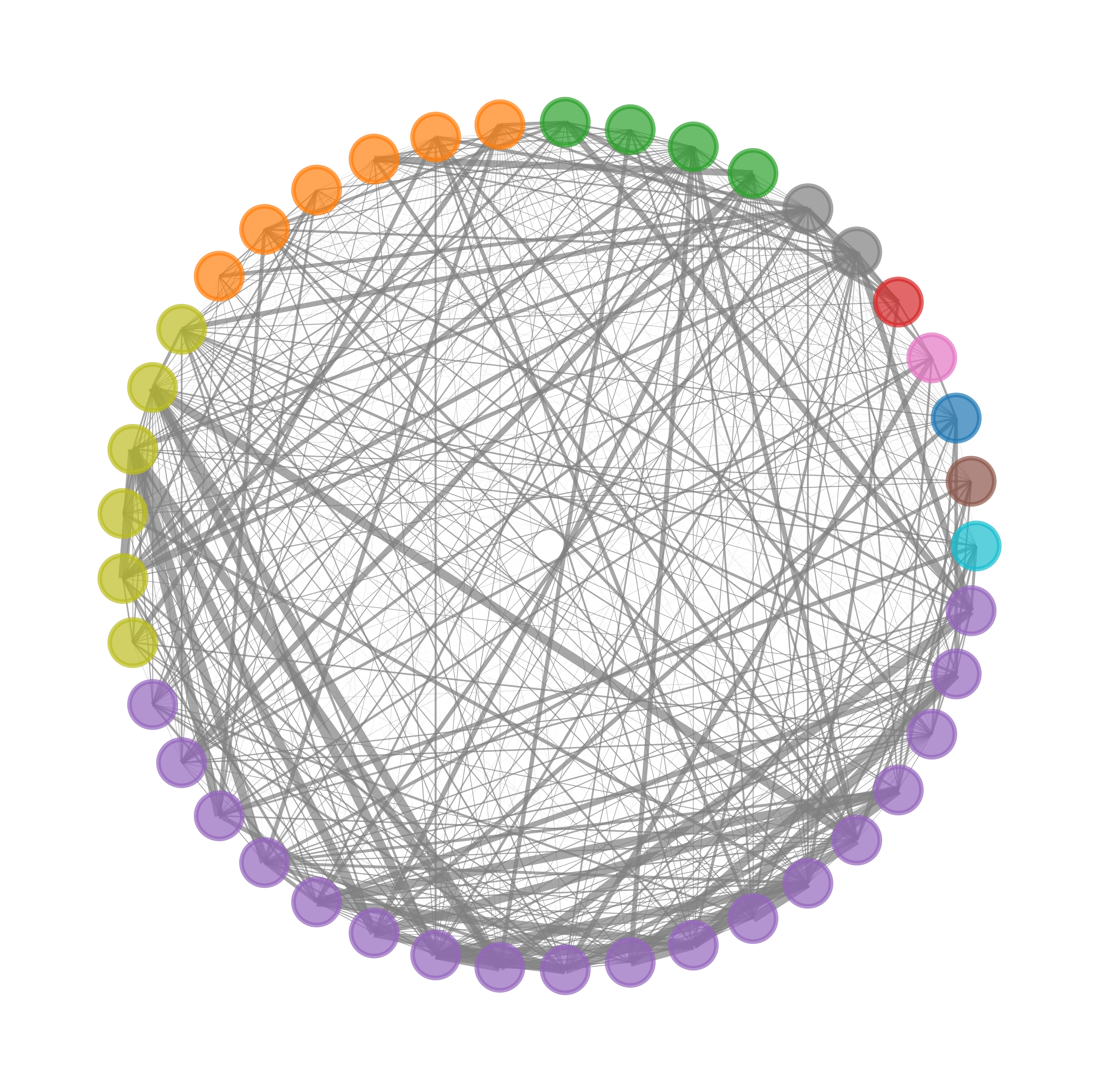} & 
        \includegraphics[width=0.18\textwidth]{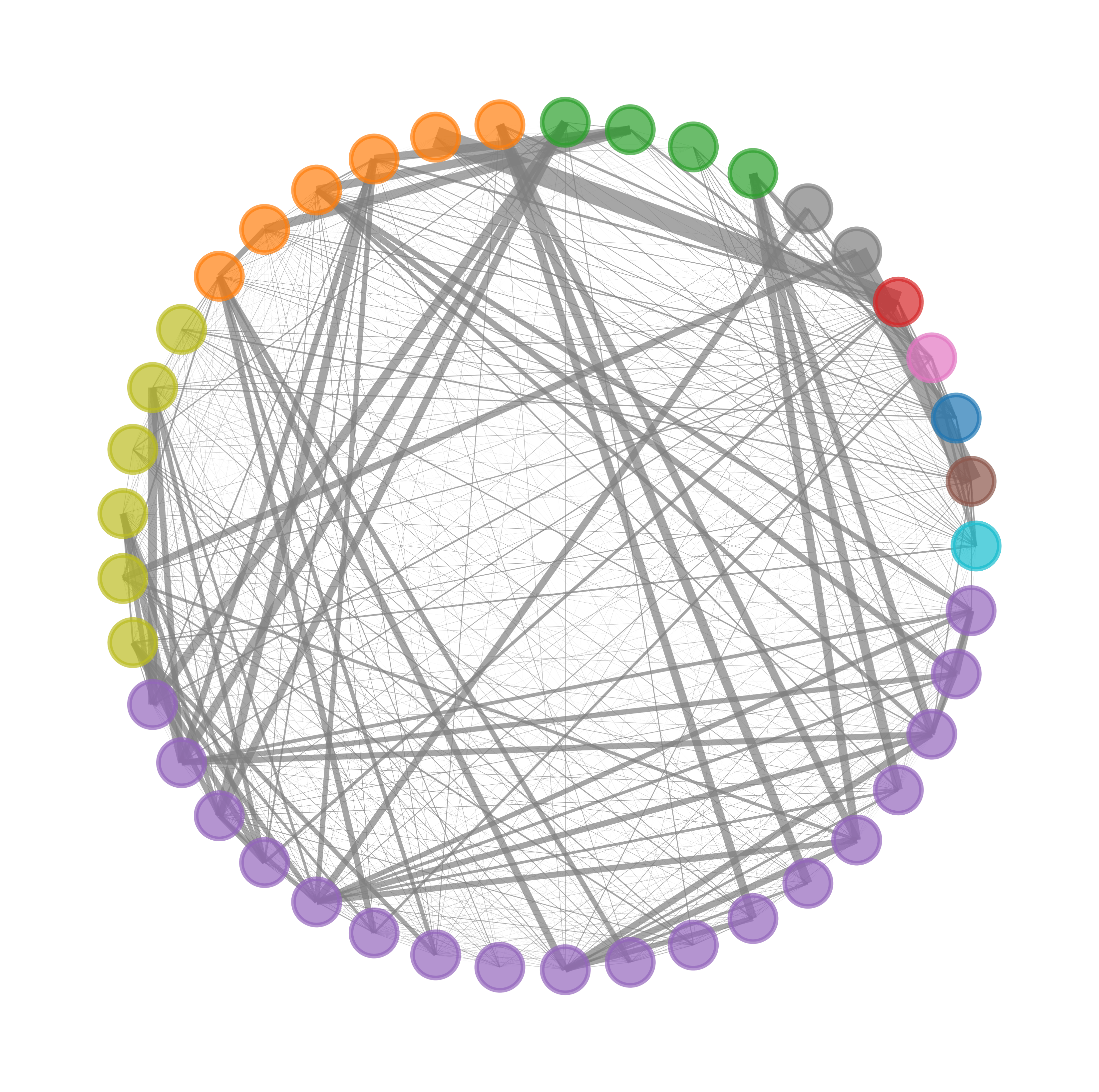} & 
        \includegraphics[width=0.18\textwidth]{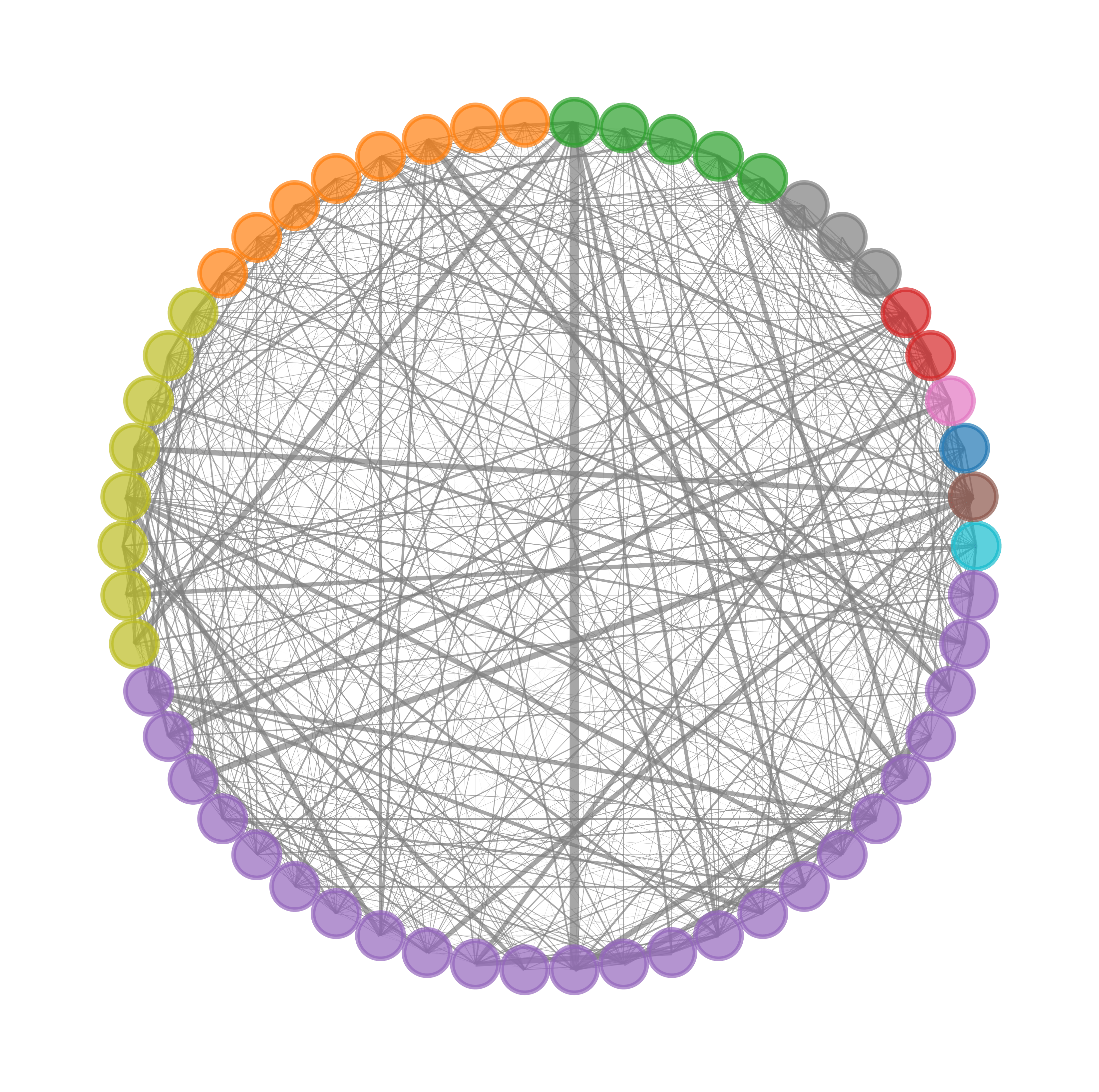} & 
        \includegraphics[width=0.18\textwidth]{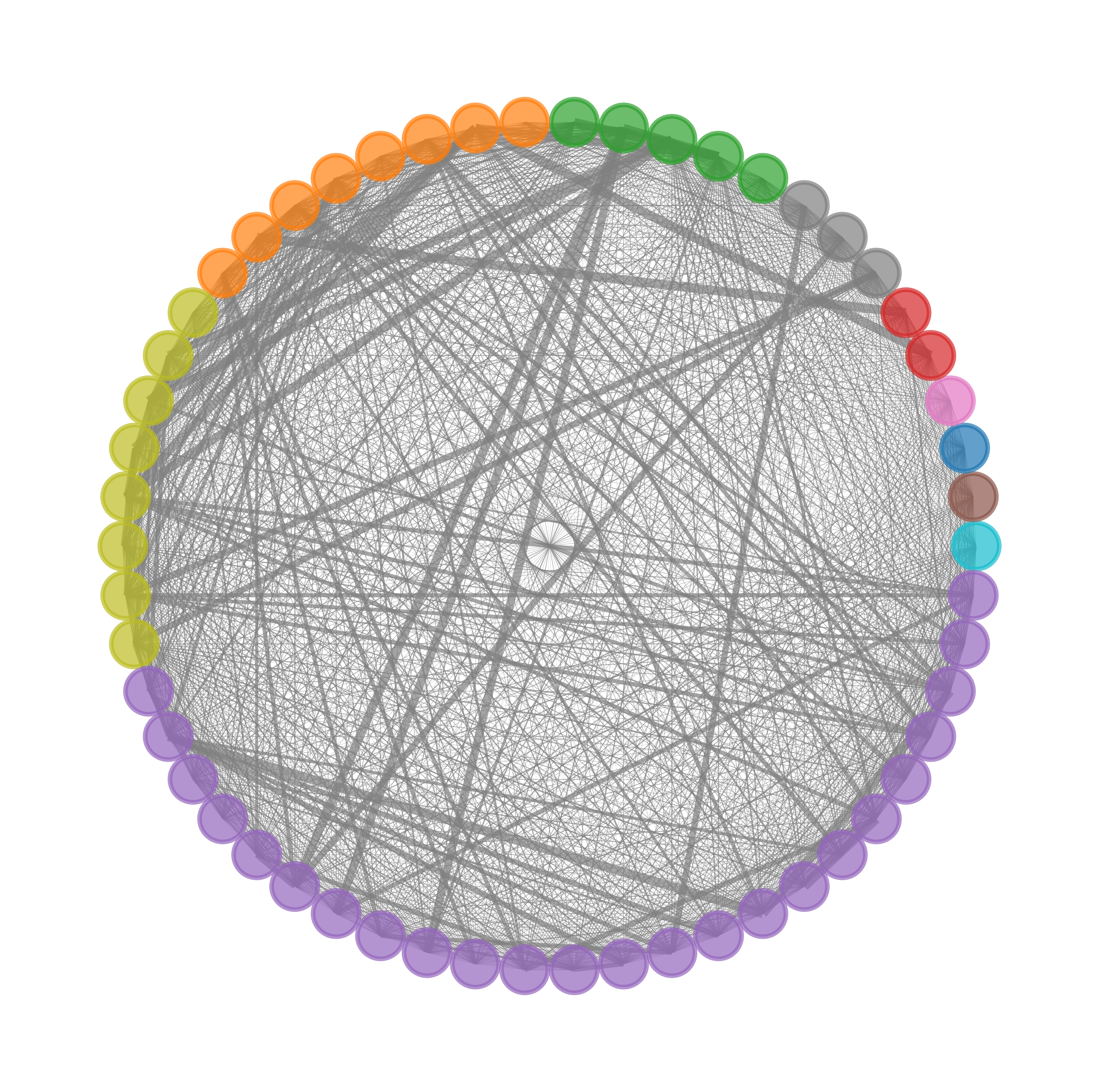} & 
        \includegraphics[width=0.18\textwidth]{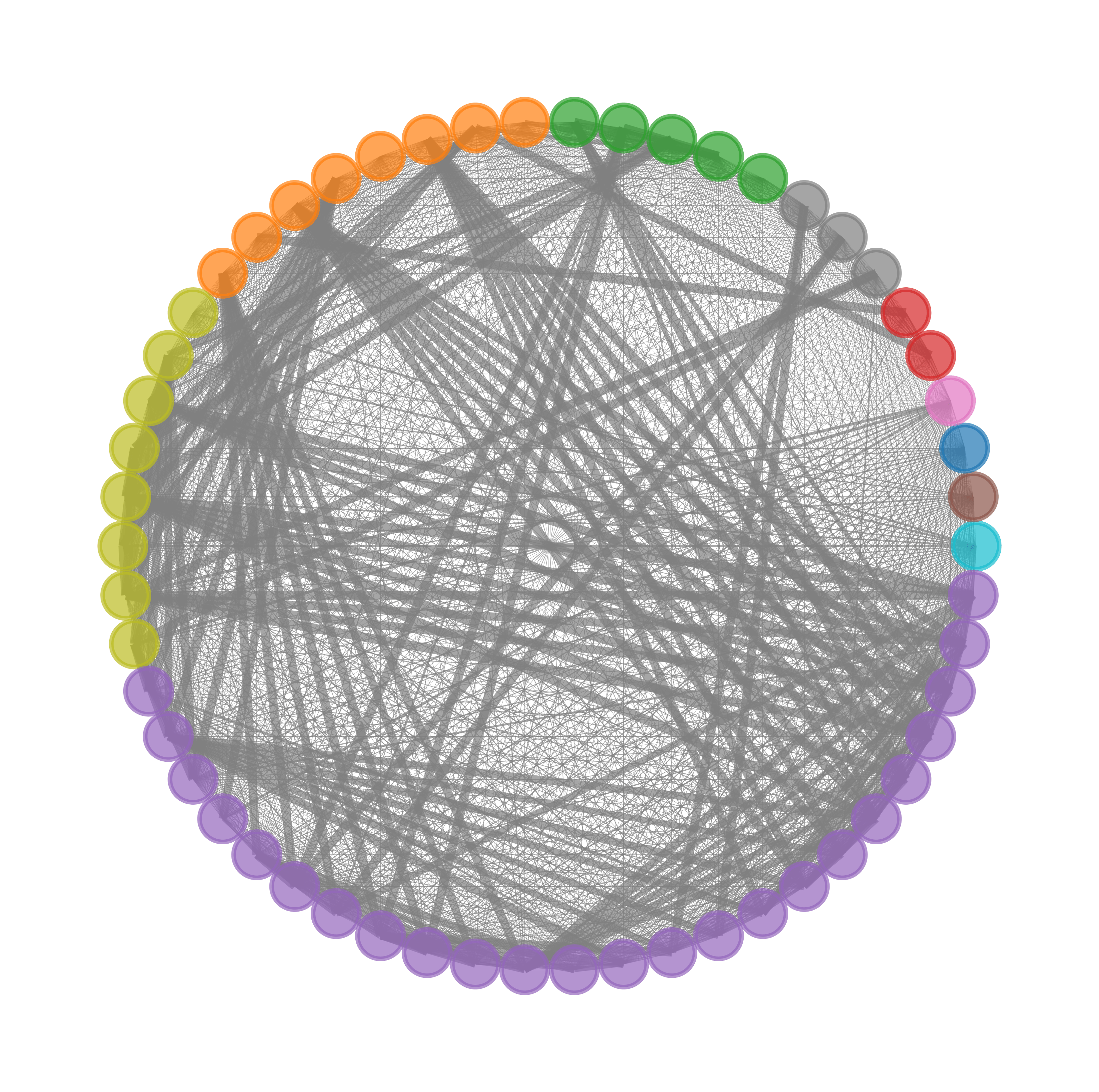} \\ [-1.5ex] 
        
        \rotatebox{90}{\makecell{Photo\\($r$=0.8\%)}} &
        \includegraphics[width=0.18\textwidth]{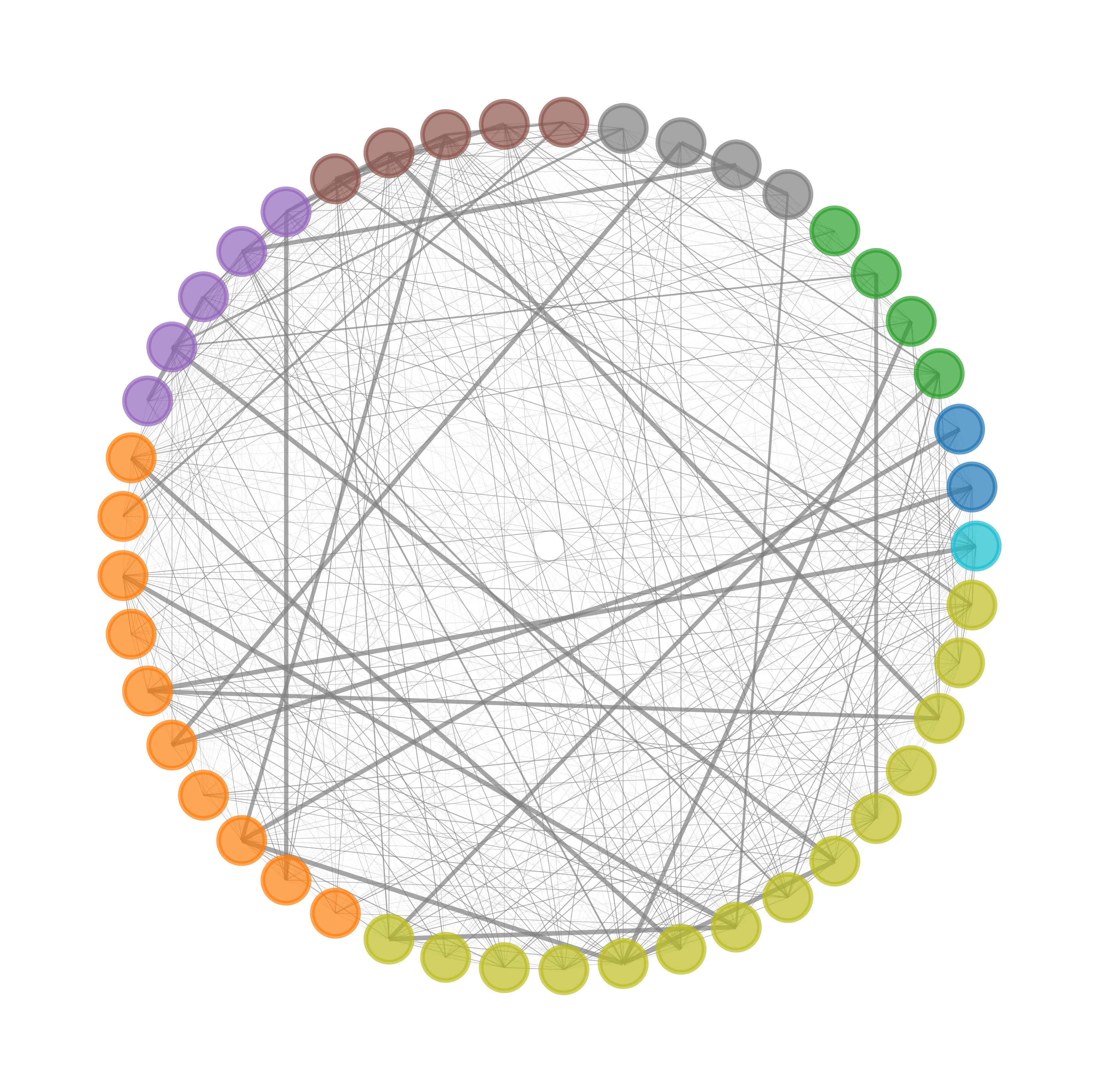} & 
        \includegraphics[width=0.18\textwidth]{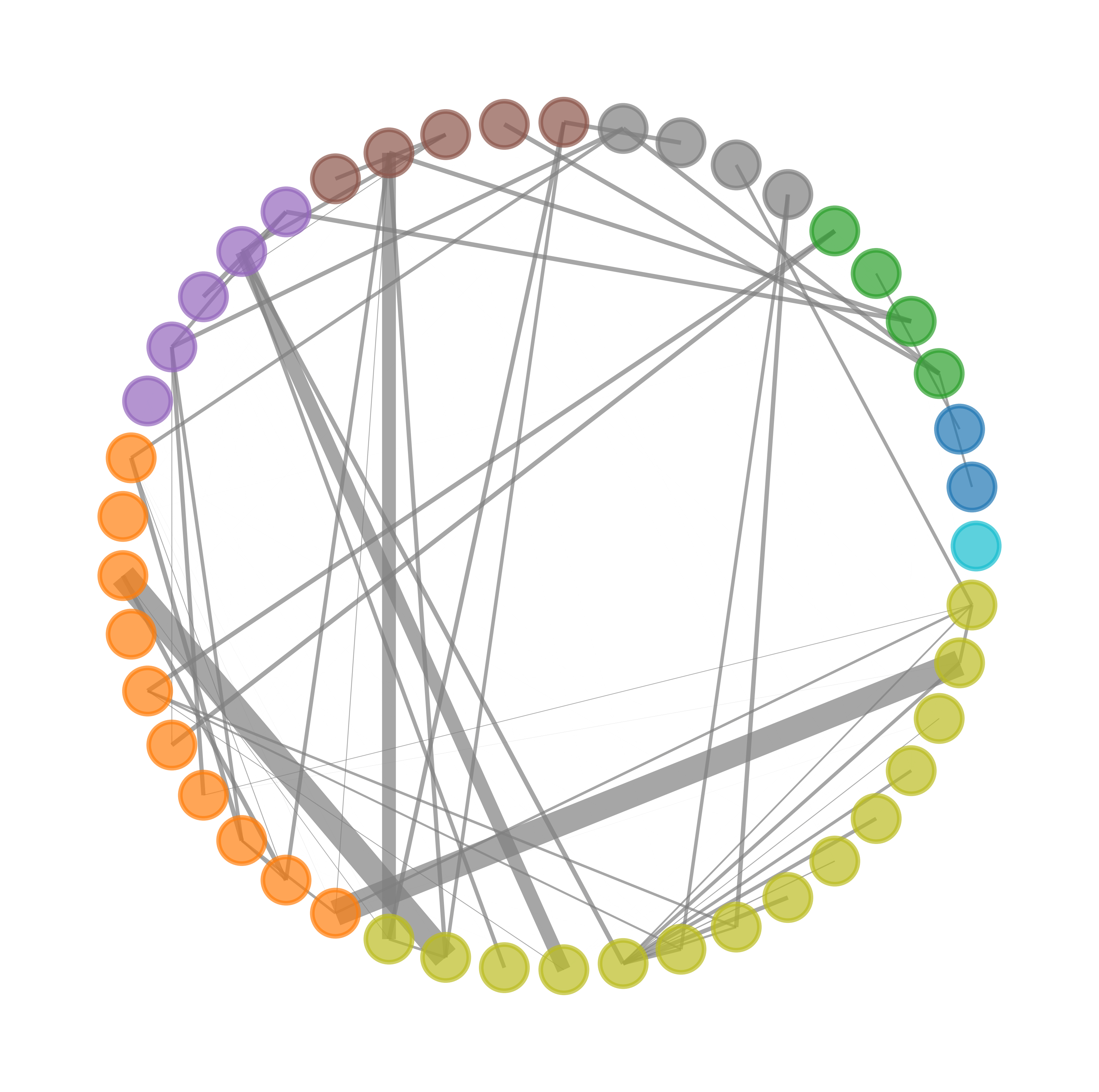} & 
        \includegraphics[width=0.18\textwidth]{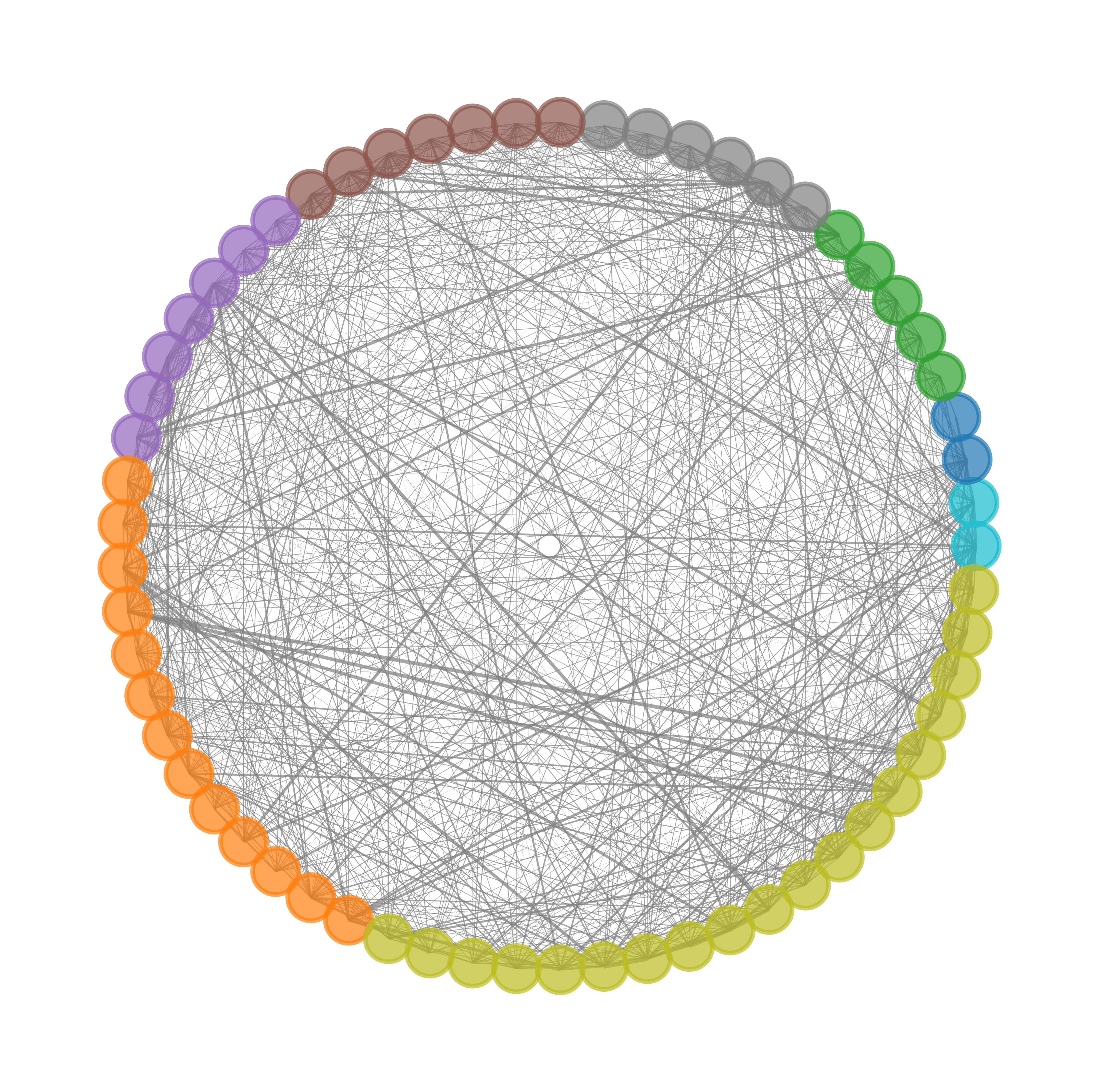} & 
        \includegraphics[width=0.18\textwidth]{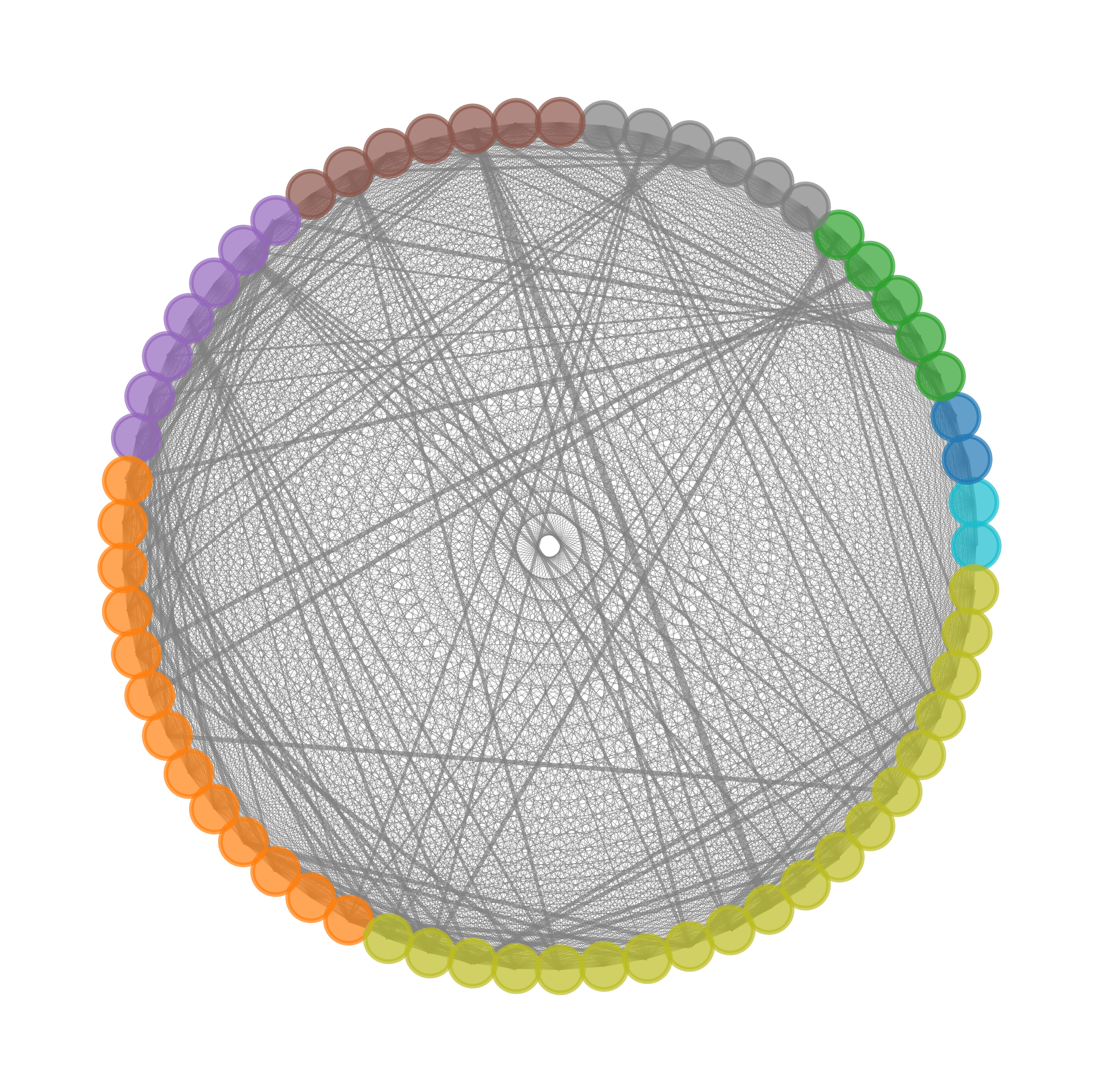} & 
        \includegraphics[width=0.18\textwidth]{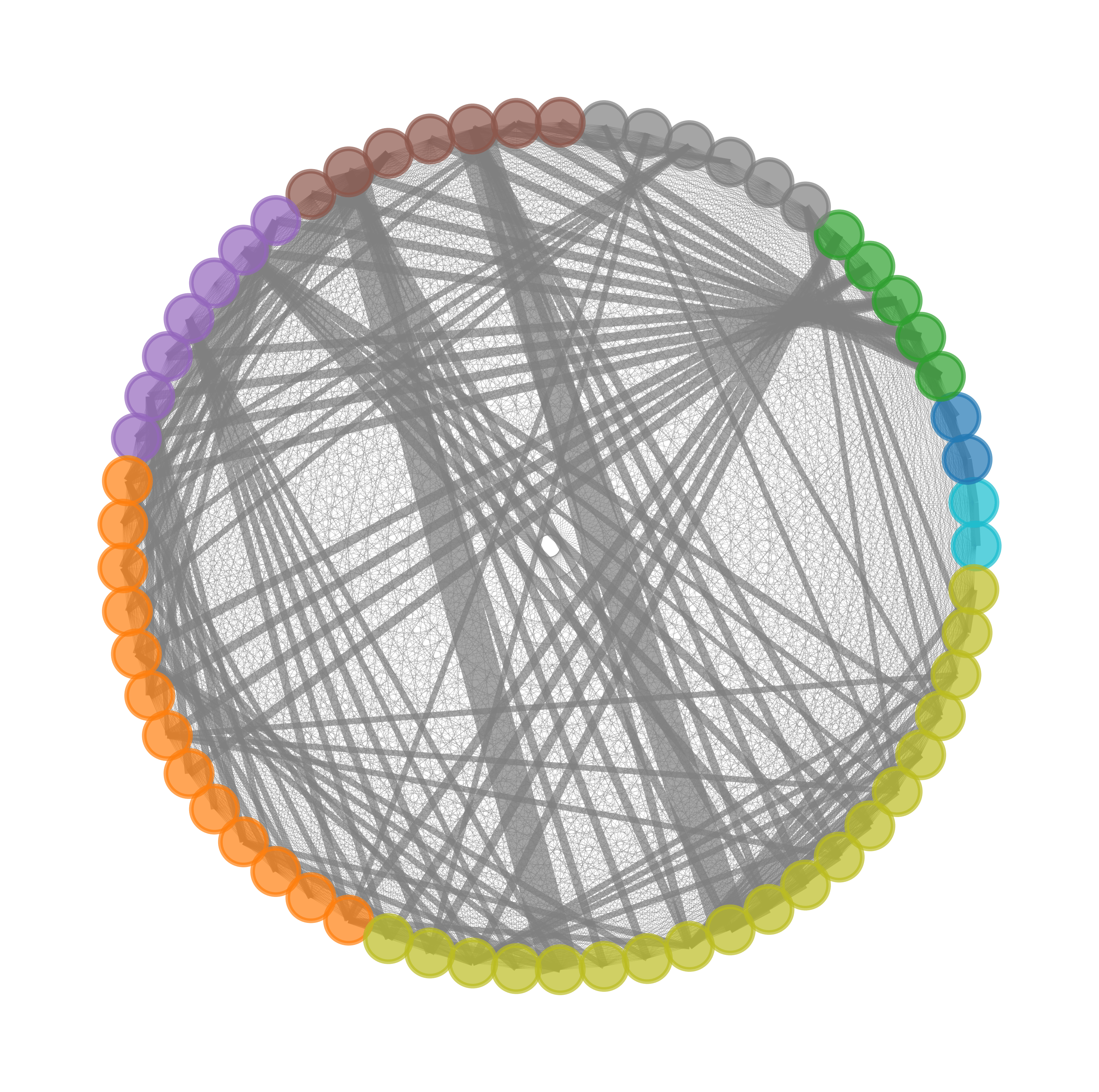} \\ [-1.5ex] 
        
        \rotatebox{90}{\makecell[t]{VAT\\($r$=1.5\%)}} & 
        \includegraphics[width=0.18\textwidth]{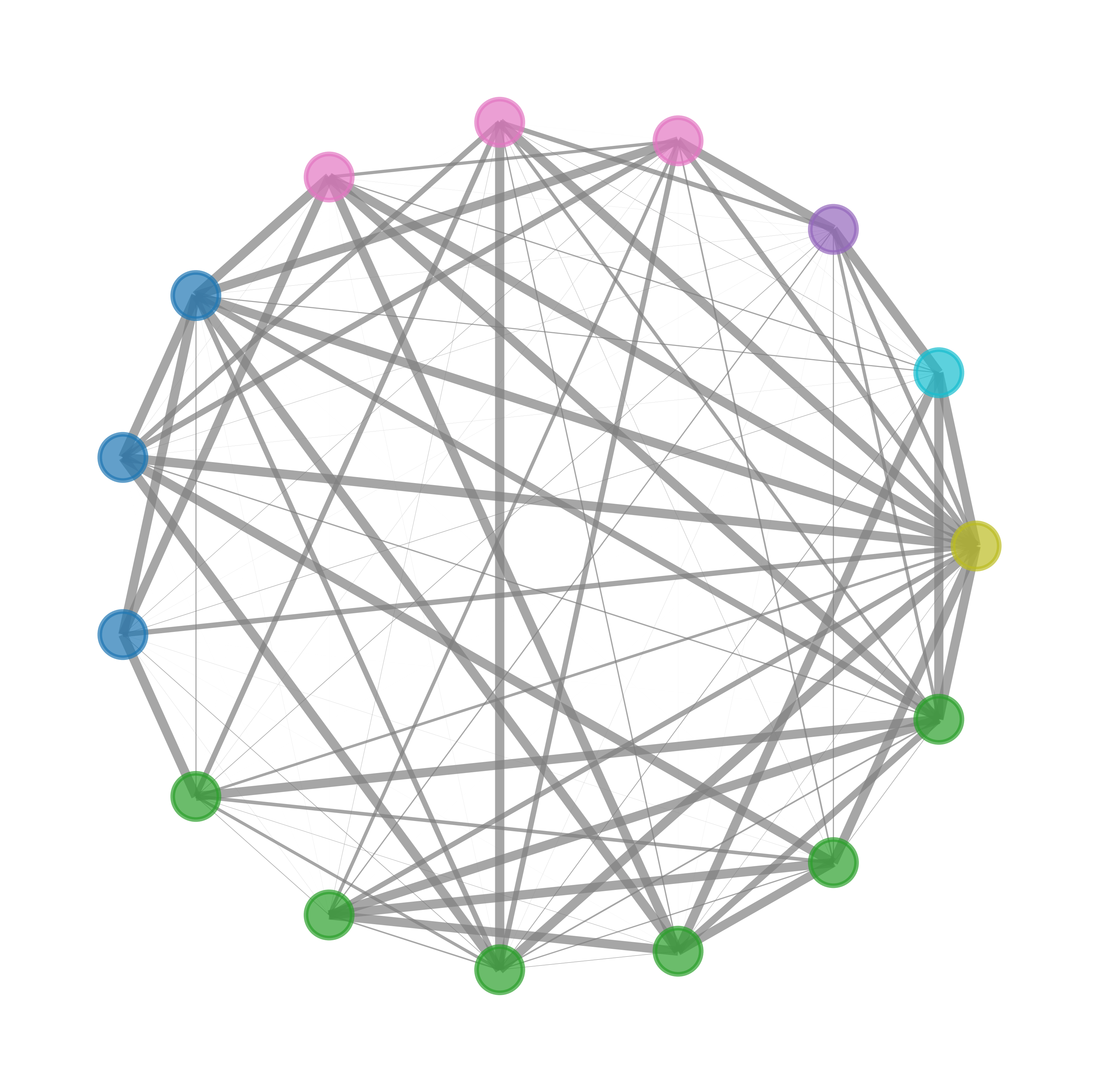} & 
        \includegraphics[width=0.18\textwidth]{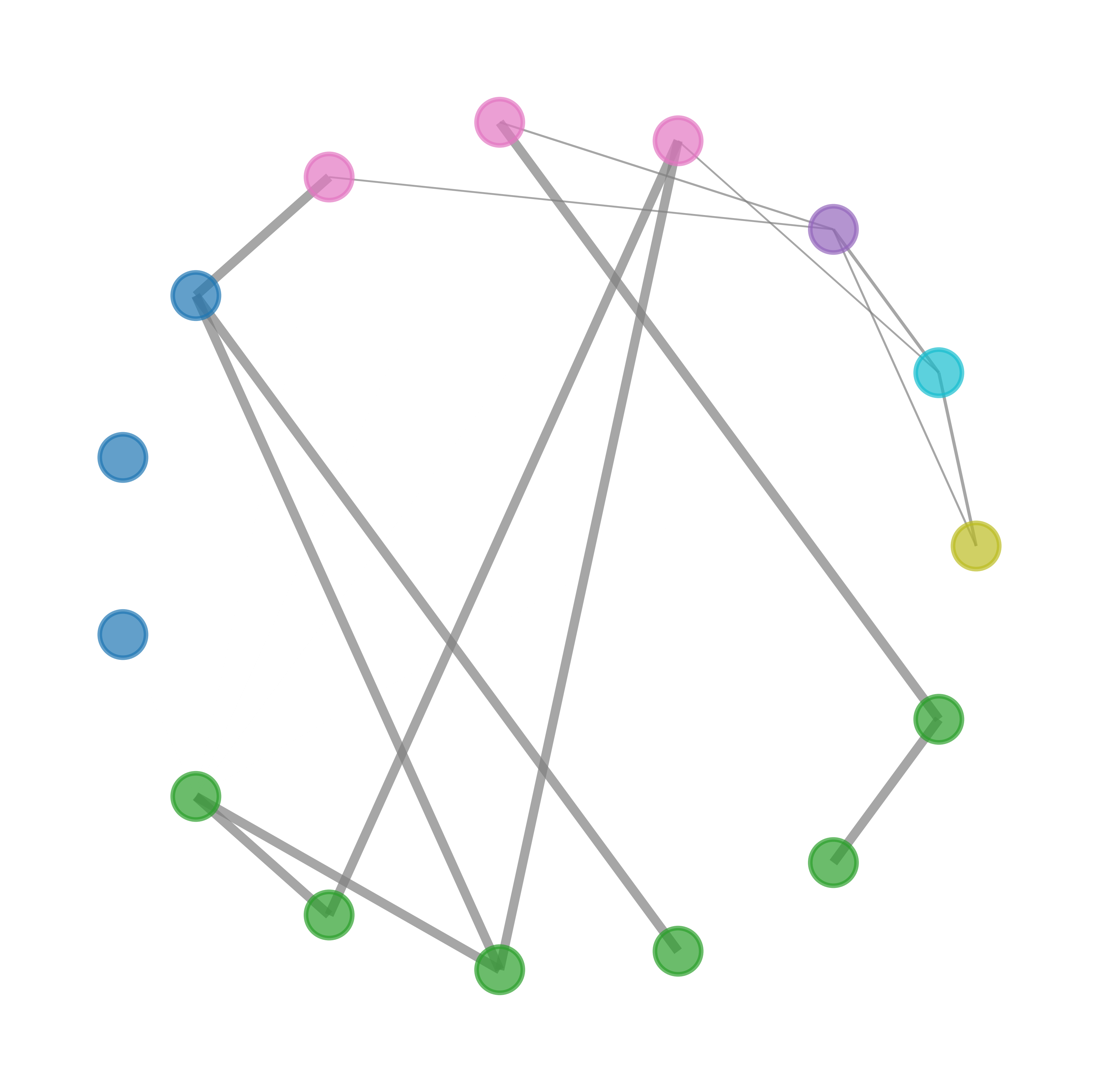} & 
        \includegraphics[width=0.18\textwidth]{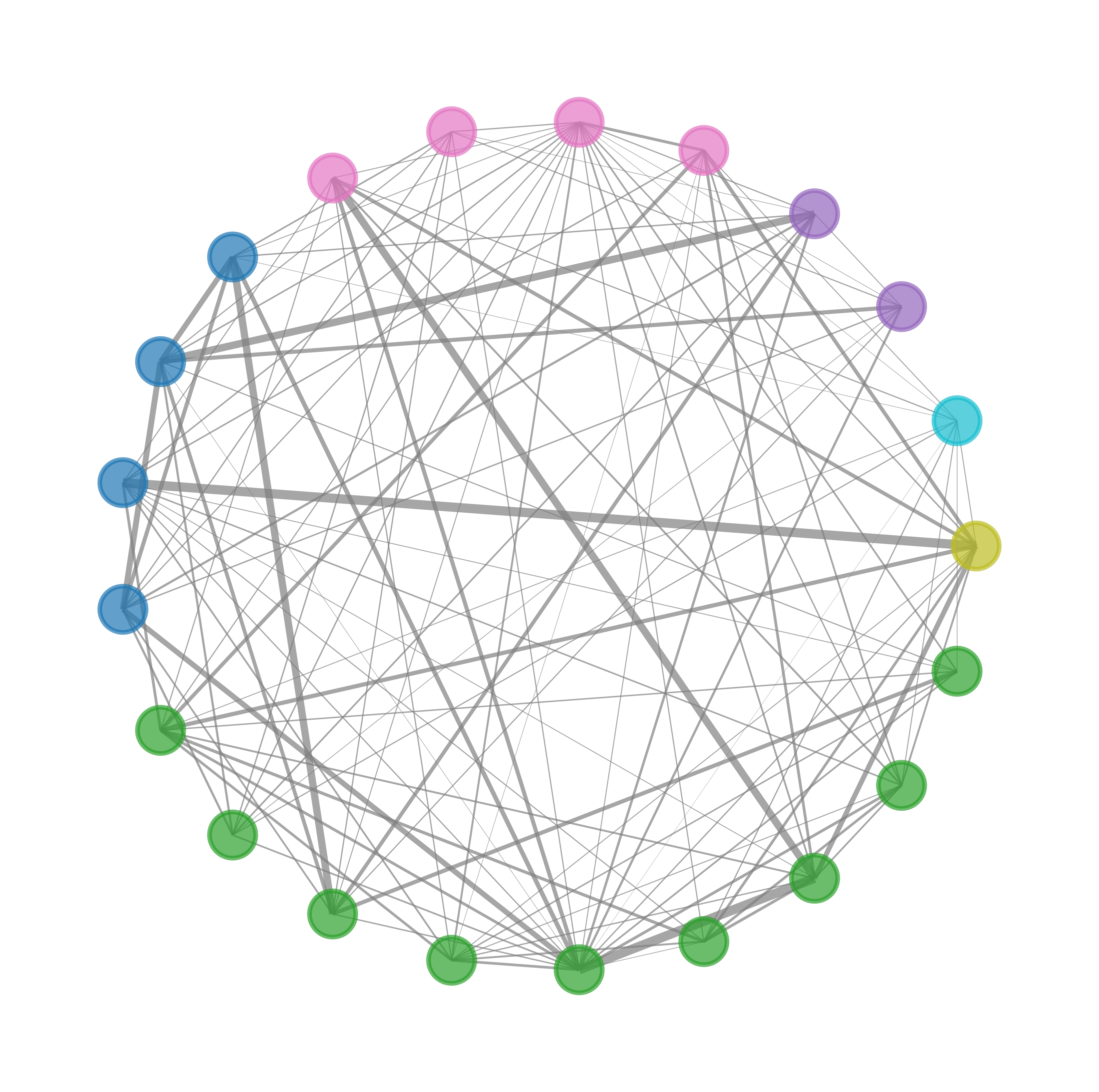} & 
        \includegraphics[width=0.18\textwidth]{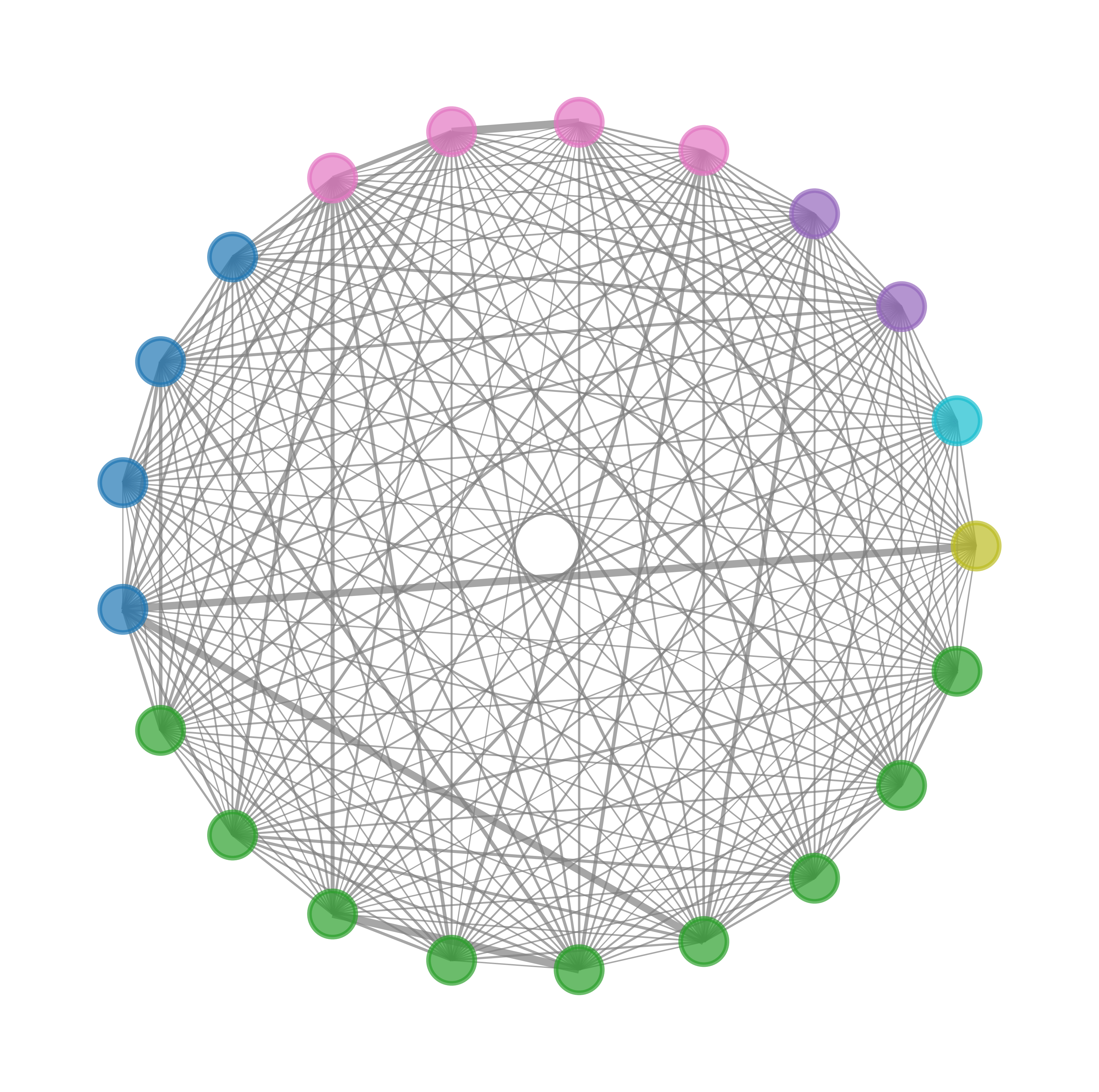} & 
        \includegraphics[width=0.18\textwidth]{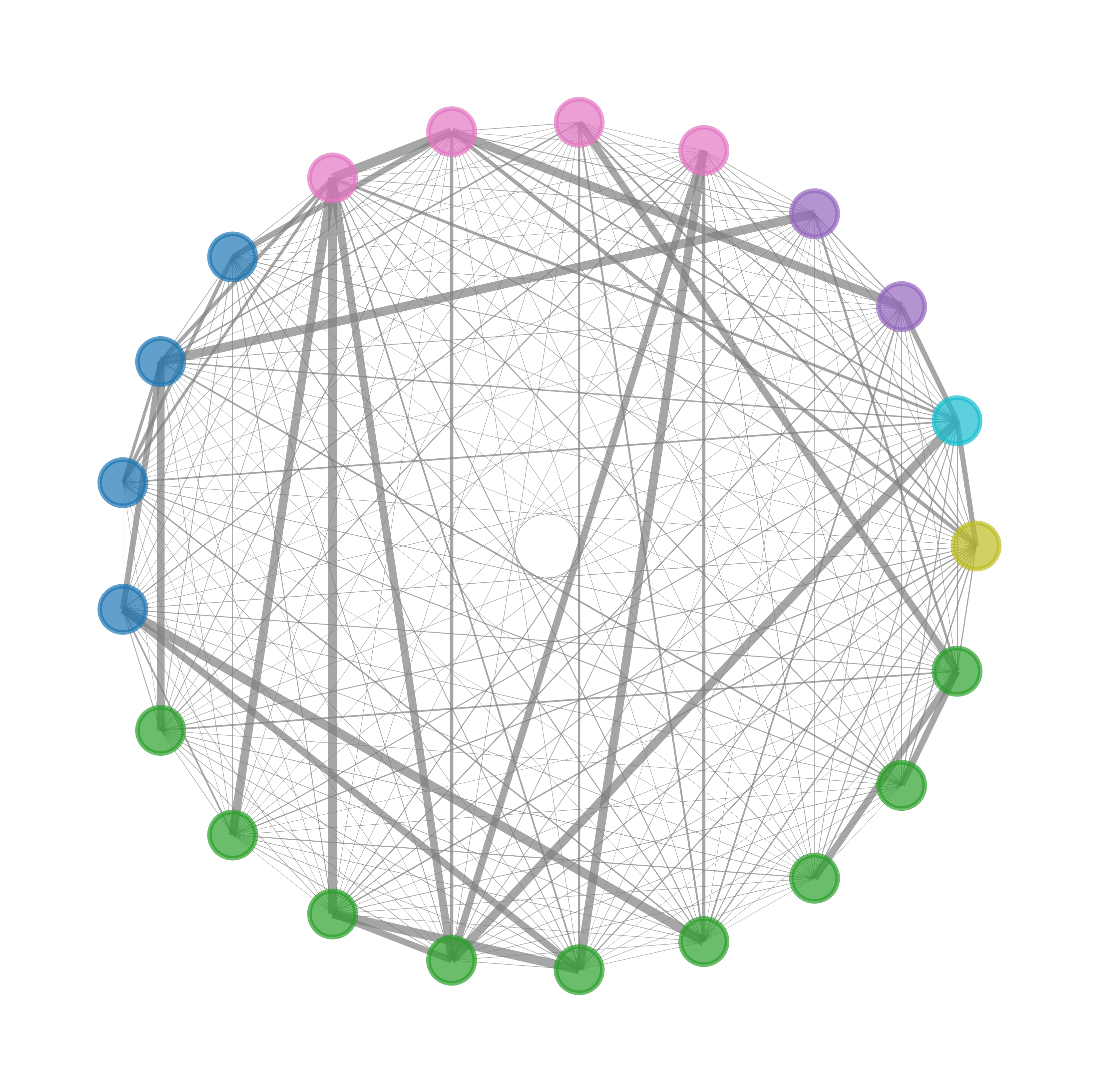} \\ [-1.5ex] 
        
        \rotatebox{90}{\makecell[c]{Automotive\\($r$=0.11\%)}} & 
        \includegraphics[width=0.17\textwidth]{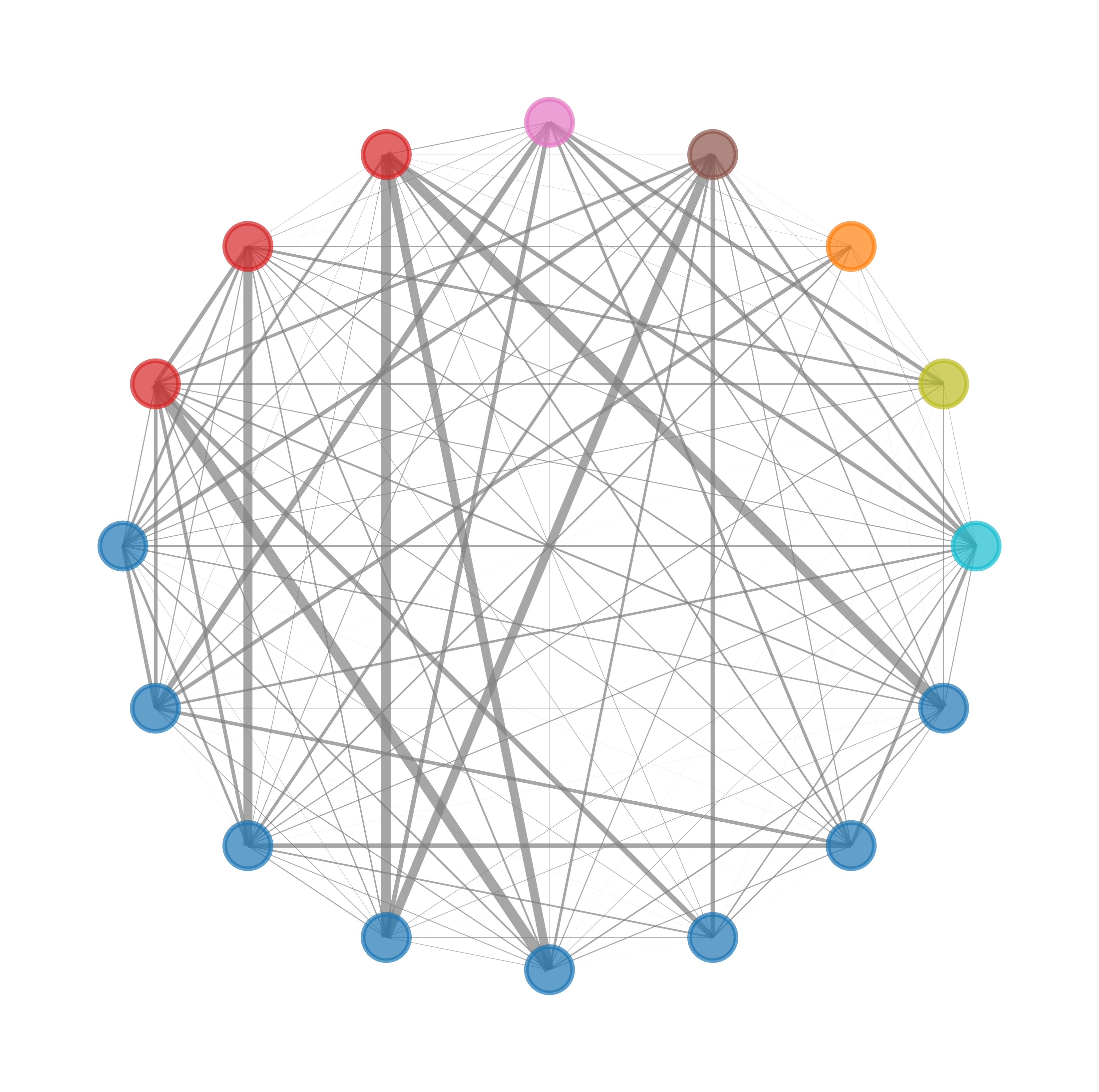} & 
        \includegraphics[width=0.17\textwidth]{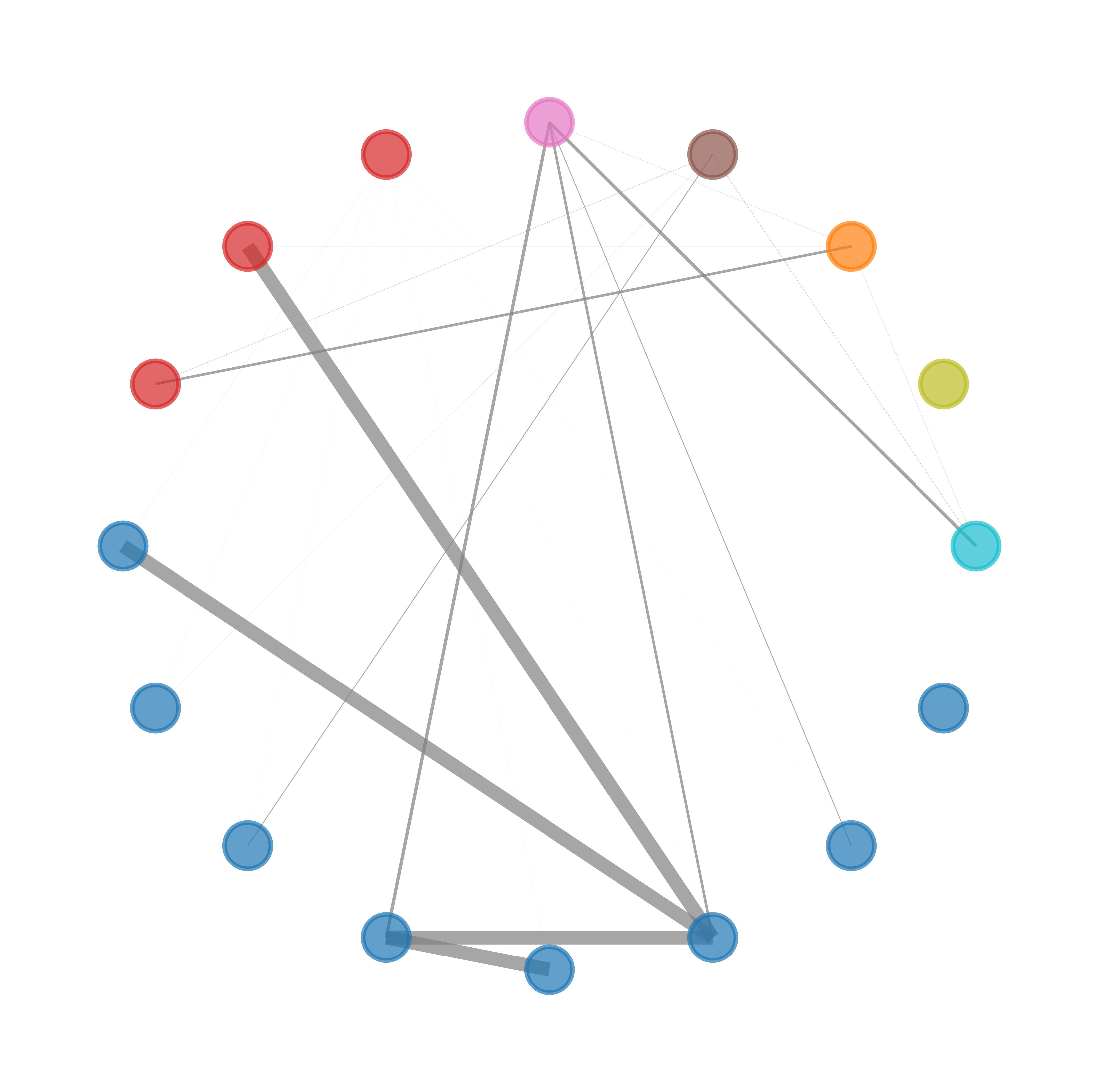} & 
        \includegraphics[width=0.17\textwidth]{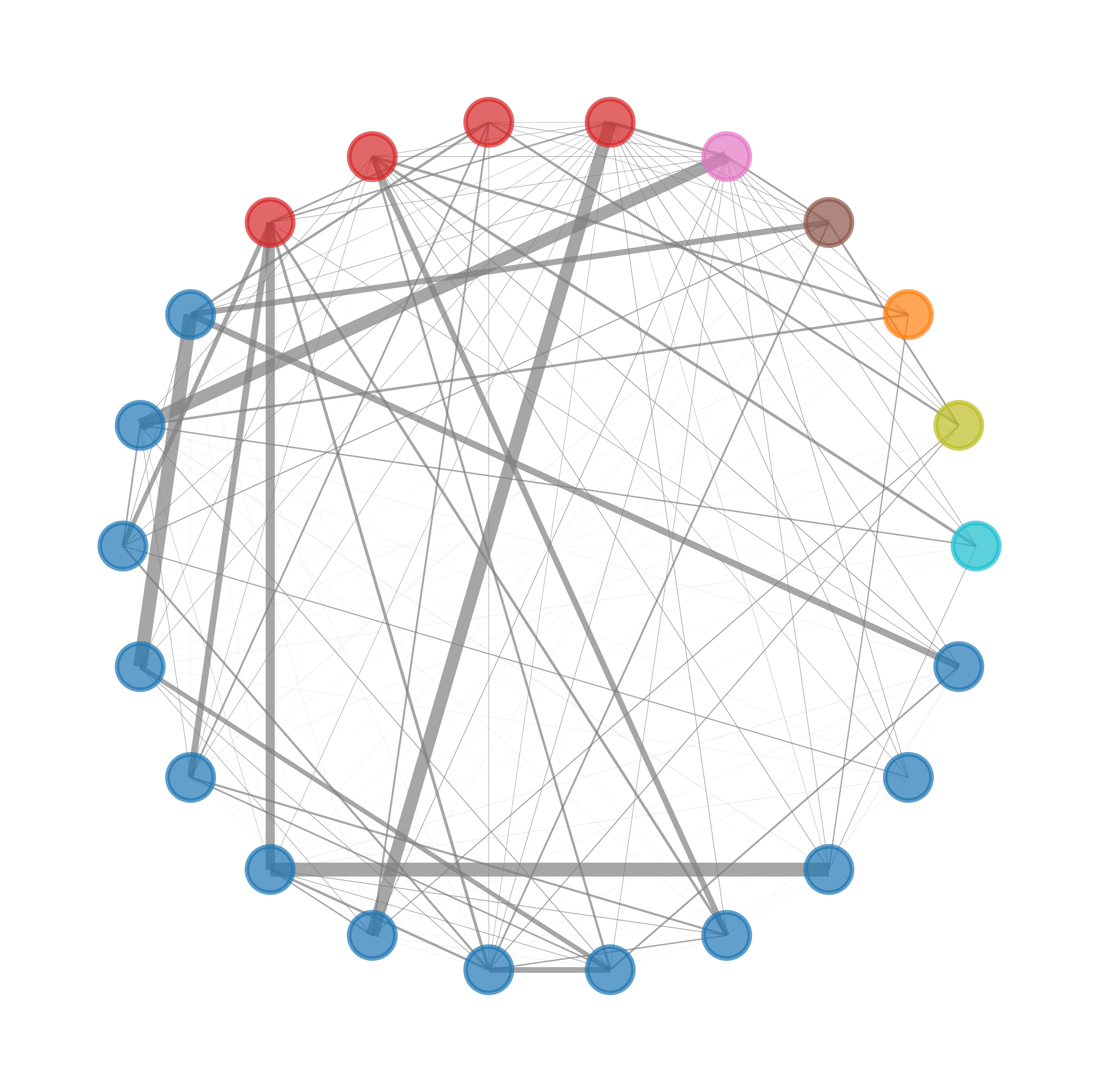} & 
        \includegraphics[width=0.17\textwidth]{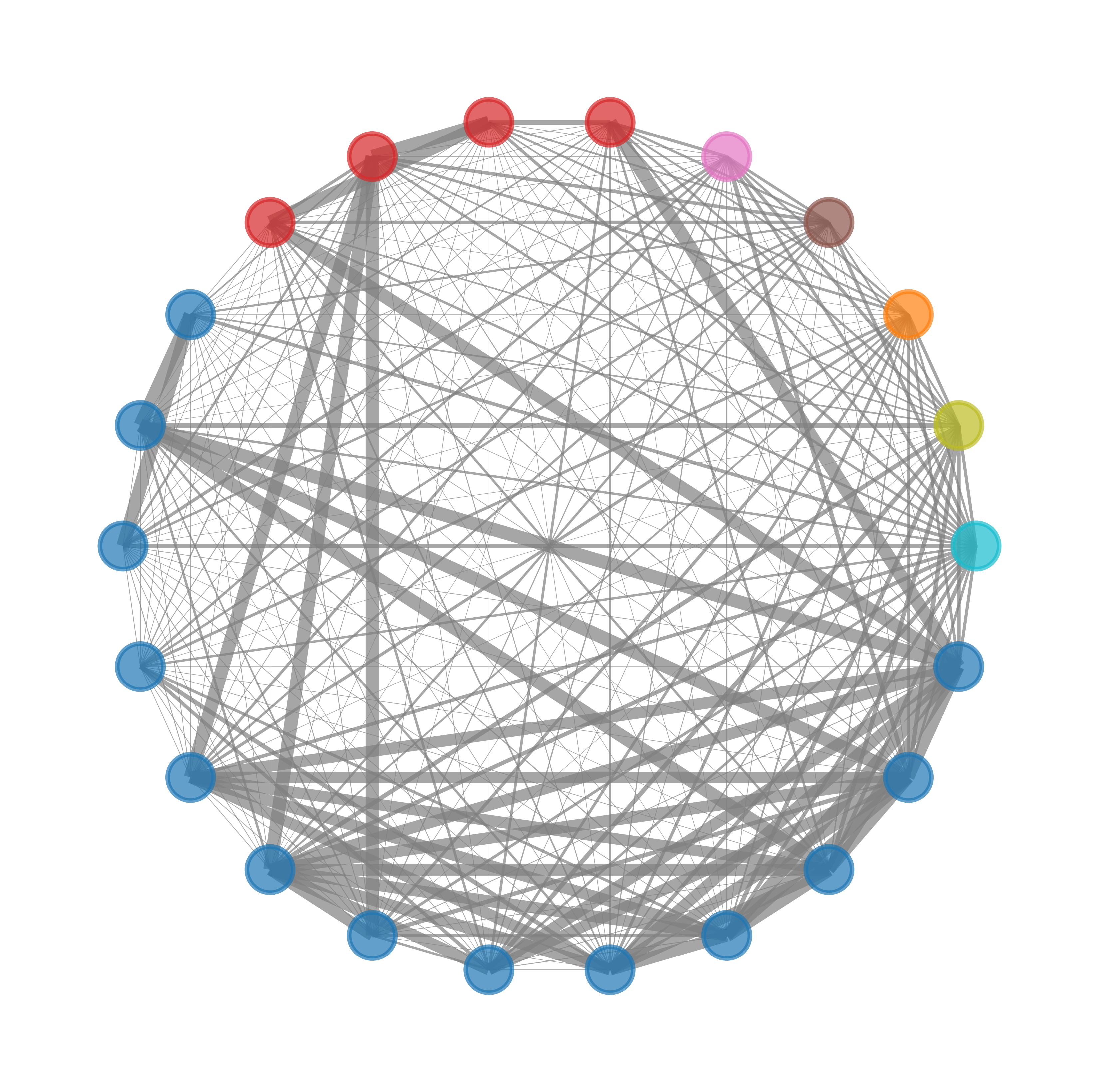} & 
        \includegraphics[width=0.17\textwidth]{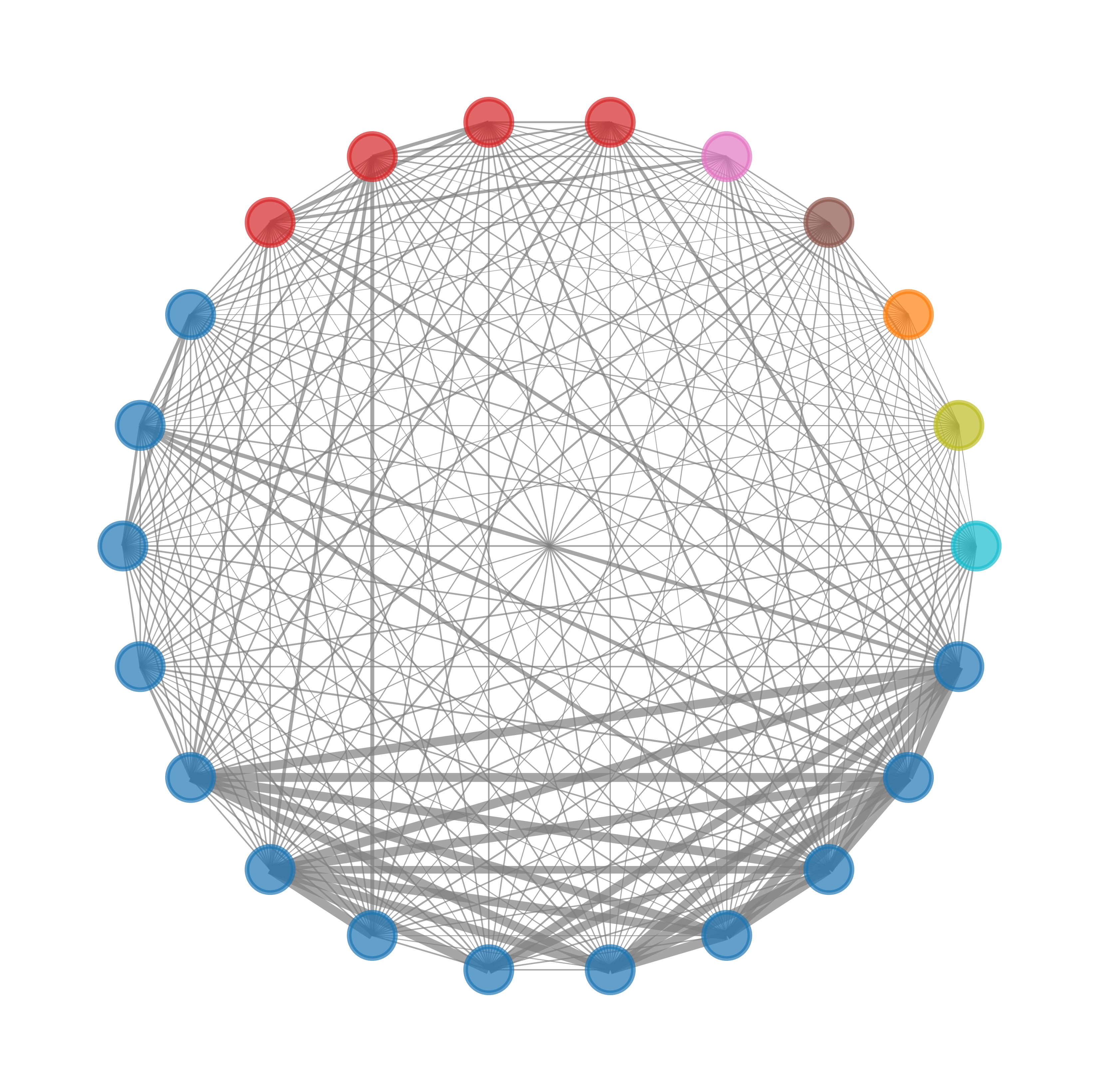} \\ [-1.5ex] 
    \end{tabular}
    \caption{
        Visualization of the condensed graphs generated by the proposed approach, HyDRO\textsuperscript{+}, and four baseline methods (GCond, SGDD, GDEM, and HyDRO) on the four datasets with the lowest reduction rate ($r$).
    }
    \label{fig:visual}
\end{figure}

\autoref{tab:statistics} compares the statistics of the condensed graphs generated by HyDRO\textsuperscript{+} with the original data. 
By drastically modifying the graph structure—such as condensing a sparse graph with only 0.05\% density into a complete graph with 100\% density, as seen in the Automotive dataset—HyDRO\textsuperscript{
+} effectively obscures the original relationships and connectivity patterns. This structural disruption makes it difficult to infer sensitive information from the condensed graph.
Additionally, \autoref{fig:visual} illustrates the topology of the condensed graphs generated by GCond, SGDD, GDEM, HyDRO and HyDRO\textsuperscript{+}. 
In these visualizations, thicker edges indicate more important connections.
As illustrated in \autoref{fig:visual}, HyDRO\textsuperscript{+} and HyDRO produce denser condensed graphs than these by GCond, SGDD, and GDEM.
Particularly, HyDRO\textsuperscript{+} generates condensed graphs with more distinct weights for different edges, highlighting key edges with very high weights. 
In contrast, HyDRO tends to produce condensed graphs with more uniform edge weights. 
These statistics and visualizations demonstrate how the condensed graphs fundamentally transform the original graph structures, enhancing privacy by reconstructing and obscuring node relationships. 
Meantimes, essential link prediction information is preserved, ensuring the utility of the condensed graphs for downstream tasks. This balance between structural transformation and task information preservation is very crucial in graph condensation.

\subsection{Privacy Preservation}
\begin{table}[h]
\renewcommand{\arraystretch}{1.5}
\centering
\caption{
    Experimental results of Link Membership Inference Attacks (LMIA) in terms of F1 score. 
    The results are shown in the format of ({\tt mean $\pm$  std dev}).
    Lower values indicate better performance. 
    The best and second-best results are highlighted in \textcolor{blue}{\textbf{blue}} and \textbf{black boldfaced}, respectively.
}
\label{tab:lmia}
\resizebox{\textwidth}{!}{%
\begin{tabular}{@{}ccccccccccclc@{}}
\toprule
\multirow{2}{*}{\textbf{Datasets}} &
  \multirow{2}{*}{\textbf{\begin{tabular}[c]{@{}c@{}} Ratio \\ (\%)\end{tabular}}} &
  \multicolumn{3}{c}{\textbf{Traditional Methods}} &
  \multicolumn{7}{c}{\textbf{Structure-based Graph Condensation Methods}} &
  \multirow{2}{*}{\textbf{\begin{tabular}[c]{@{}c@{}}Whole \\ Dataset\end{tabular}}} \\ \cmidrule(lr){3-5} \cmidrule(lr){6-12}
 &
   &
  Random &
  Herding &
  KCenter &
  GCond &
  SGDD &
  DosCond &
  MSGC &
  GDEM &
  HyDRO &
  HyDRO\textsuperscript{$\mathbf{+}$} & \\ \midrule
\multirow{3}{*}{Computers} &
  0.40\% &
  41.81±0.34 &
  41.13±0.34 &
  44.67±1.70 &
  \textcolor{blue}{\bf{40.02±0.07}} &
  54.68±0.26 &
  \bf{40.76±0.22} &
  44.95±2.27 &
  41.31±1.69 &
  54.27±0.40 &
  53.23±0.23 &
  \multirow{3}{*}{56.98±0.26} \\ \cline{3-12}
  &
  0.80\% &
  42.12±0.78 &
  41.81±0.72 &
  47.24±2.43 &
  \bf{40.36±0.16} &
  53.02±3.09 &
  49.32±0.69 &
  42.29±2.20 &
  \textcolor{blue}{\bf{40.01±0.01}} &
  54.89±0.17 &
  54.42±0.36 & \\ \cline{3-12}
  &
  1.60\% &
  42.61±0.79 &
  42.15±0.53 &
  49.30±3.01 &
  50.08±0.68 &
  50.14±0.03 &
  \bf{40.67±0.44} &
  42.90±4.74 &
  \textcolor{blue}{\bf{40.21±0.24}} &
  53.42±1.86 &
  52.59±2.17 & \\ \midrule
\multirow{3}{*}{Photo} &
  0.80\% &
  42.93±1.31 &
  \bf{41.46±0.98} &
  51.64±1.57 &
  48.79±1.84 &
  56.09±1.18 &
  49.91±1.93 &
  50.66±4.31 &
  \textcolor{blue}{\bf{40.01±0.01}} &
  56.67±0.98 &
  \multicolumn{1}{l}{55.49±1.72} &
  \multirow{3}{*}{60.70±1.40} \\ \cline{3-12}
 &
  1.60\% &
  45.29±1.68 &
  \bf{43.07±0.92} &
  53.88±1.38 &
  52.51±1.72 &
  57.26±1.87 &
  44.74±1.39 &
  45.62±3.54 &
  \textcolor{blue}{\bf{40.03±0.04}} &
  56.97±0.78 &
  \multicolumn{1}{l}{56.41±0.37} & \\ \cline{3-12}
   &
  4.00\% &
  51.21±1.38 &
  47.49±1.12 &
  55.91±0.98 &
  51.79±1.37 &
  57.18±2.38 &
  \bf{44.42±1.34} &
  50.48±5.03 &
  \textcolor{blue}{\bf{43.63±3.49}} &
  55.49±3.93 &
  \multicolumn{1}{l}{56.41±0.41} &
   \\ \midrule
\multirow{3}{*}{VAT} &
  1.50\% &
  \bf{40.27±0.13} &
  40.42±0.11 &
  40.59±0.35 &
  42.62±0.63 &
  45.81±0.58 &
  41.36±0.14 &
  43.01±0.69 &
  \textcolor{blue}{\bf{40.12±0.02}} &
  45.63±0.49 &
  \multicolumn{1}{l}{45.80±0.28} &
  \multirow{3}{*}{46.80±0.31} \\ \cline{3-12}
  &
  3.00\% &
  41.27±0.22 &
  41.74±0.22 &
  41.38±0.44 &
  42.25±0.25 &
  43.36±1.05 &
  41.96±0.11 &
  43.90±1.50 &
  \textcolor{blue}{\bf{40.18±0.12}} &
  41.77±0.39 &
  \bf{41.22±0.65} &
   \\ \cline{3-13}
  &
  6.00\% &
  41.79±0.28 &
  42.04±0.26 &
  \bf{41.37±0.23} &
  43.39±0.61 &
  45.76±0.35 &
  45.41±0.39 &
  44.48±0.78 &
  \textcolor{blue}{\bf{40.24±0.15}} &
  43.89±0.49 &
  43.44±0.68 & \\ \midrule
\multirow{3}{*}{Automotive} &
  0.11\% &
  45.84±0.68 &
  45.15±0.82 &
  43.32±0.11 &
  48.10±0.27 &
  47.46±0.30 &
  \bf{42.12±1.15} &
  48.48±0.41 &
  44.20±0.98 &
  48.54±0.37 &
  \textcolor{blue}{\bf{40.10±1.47}} &
  \multirow{3}{*}{50.75±0.28} \\ \cline{3-12}
  &
  0.56\% &
  49.32±0.55 &
  47.55±0.81 &
  47.99±0.33 &
  48.05±0.20 &
  48.78±0.69 &
  46.98±0.99 &
  \textcolor{blue}{\bf{44.80±1.30}} &
  \bf{45.33±1.62} &
  47.84±0.43 &
  46.21±0.59 & \\ \cline{3-12}
   &
  1.12\% &
  49.90±0.71 &
  49.61±0.24 &
  47.50±0.50 &
  47.16±0.45 &
  48.30±0.29 &
  47.86±0.56 &
  46.70±0.62 &
  \textcolor{blue}{\bf{45.70±1.04}} &
  48.17±0.24 &
  \bf{46.52±0.50} &
   \\ \bottomrule
\end{tabular}%
}
\end{table}
\begin{table}[h]
\renewcommand{\arraystretch}{1.5}
\centering
\caption{
    Experimental results for the accuracy on Membership Inference Attacks (MIA) on nodes. 
    The results are shown in the format of ({\tt mean $\pm$  std dev}).
    Lower values indicate better performance. 
    The best and second-best results are highlighted in \textcolor{blue}{\textbf{blue}} and \textbf{black boldfaced}, respectively.
}
\label{tab:mia}
\resizebox{\textwidth}{!}{%
\begin{tabular}{@{}ccccccccccccc@{}}
\toprule
\multirow{2}{*}{\textbf{Datasets}} &
  \multirow{2}{*}{\textbf{\begin{tabular}[c]{@{}c@{}}Ratio \\ (\%)\end{tabular}}} &
  \multicolumn{3}{c}{\textbf{Traditional Methods}} &
  \multicolumn{7}{c}{\textbf{Structure-based Graph Condensation Methods}} &
  \multirow{2}{*}{\textbf{\begin{tabular}[c]{@{}c@{}}Whole \\ Dataset\end{tabular}}} \\ \cmidrule(lr){3-5} \cmidrule(lr){6-12}
 &
   &
  Random &
  Herding &
  KCenter &
  GCond &
  SGDD &
  DosCond &
  MSGC &
  GDEM &
  HyDRO &
  HyDRO\textsuperscript{$\mathbf{+}$} &
   \\ \hline
\multirow{3}{*}{Computers} &
  0.40\% &
  50.46±0.15 &
  50.32±0.06 &
  50.53±0.10 &
  \bf{50.02±0.01} &
  50.09±0.08 &
  50.22±0.06 &
  50.38±0.09 &
  50.23±0.07 &
  50.17±0.05 &
  \textcolor{blue}{\bf{50.00±0.00}} &
  \multirow{3}{*}{51.34±0.12} \\ \cline{3-12}
 &
  0.80\% &
  50.22±0.19 &
  50.31±0.05 &
  50.50±0.15 &
  50.12±0.01 &
  \textcolor{blue}{\bf{50.01±0.01}} &
  50.46±0.07 &
  50.26±0.01 &
  50.28±0.08 &
  50.12±0.04 &
  \bf{50.11±0.02} & \\ \cline{3-12}
  &
  1.60\% &
  50.31±0.09 &
  50.27±0.08 &
  50.55±0.07 &
  50.11±0.03 &
  50.14±0.03 &
  50.18±0.04 &
  50.25±0.08 &
  \textcolor{blue}{\bf{50.00±0.01}} &
  \bf{50.10±0.05} &
  50.12±0.01 &
   \\ \midrule
\multirow{3}{*}{Photo} &
  0.80\% &
  50.54±0.99 &
  51.11±0.16 &
  50.53±0.07 &
  50.30±0.13 &
  \bf{50.22±0.08} &
  \textcolor{blue}{\bf{51.16±0.14}} &
  50.42±0.16 &
  50.34±0.25 &
  50.30±0.09 &
  50.40±0.15 &
  \multirow{3}{*}{50.56±0.22} \\ \cline{3-12}
 &
  1.60\% &
  50.36±0.08 &
  50.95±0.15 &
  50.67±0.15 &
  \bf{50.23±0.05} &
  50.26±0.11 &
  51.06±0.23 &
  50.87±0.12 &
  \textcolor{blue}{\bf{50.20±0.08}} &
  50.52±0.11 &
  50.36±0.08 & \\ \cline{3-12}
 &
  4.00\% &
  50.48±0.03 &
  50.77±0.18 &
  50.71±0.17 &
  50.31±0.36 &
  50.30±0.14 &
  50.59±0.16 &
  50.48±0.16 &
  \textcolor{blue}{\bf{50.13±0.10}} &
  50.30±0.17 &
  \bf{50.22±0.07} &
   \\ \midrule
\multirow{3}{*}{VAT} &
  1.50\% &
  53.71±0.37 &
  52.88±0.85 &
  53.89±0.38 &
  52.92±0.81 &
  \textcolor{blue}{\bf{52.67±1.22}} &
  \bf{52.70±0.36} &
  53.33±0.28 &
  53.79±0.46 &
  54.37±0.61 &
  53.83±0.47 &
  \multirow{3}{*}{53.18±0.89} \\ \cline{3-12}
  &
  3.00\% &
  54.52±0.64 &
  53.74±0.42 &
  53.08±0.84 &
  52.57±0.69 &
  \bf{52.22±0.59} &
  \textcolor{blue}{\bf{52.04±0.38}} &
  53.87±0.49 &
  53.59±0.66 &
  52.81±0.38 &
  52.39±0.59 & \\ \cline{3-12}
  &
  6.00\% &
  53.86±0.73 &
  \bf{52.37±0.78} &
  52.65±0.50 &
  53.55±0.76 &
  53.55±0.76 &
  \textcolor{blue}{\bf{51.73±0.79}} &
  53.69±0.54 &
  54.72±0.49 &
  52.87±0.35 &
  52.96±0.32 & \\ \midrule
\multirow{3}{*}{Automotive} &
  0.11\% &
  54.02±0.30 &
  \bf{53.42±0.62} &
  53.66±0.23 &
  53.79±0.32 &
  54.09±0.37 &
  54.39±0.25 &
  53.78±0.52 &
  \textcolor{blue}{\bf{53.08±0.61}} &
  53.97±0.43 &
  53.81±0.29 &
  \multirow{3}{*}{53.12±0.44} \\ \cline{3-12}
  &
  0.56\% &
  53.59±0.13 &
  53.86±0.25 &
  53.86±0.13 &
  53.56±0.42 &
  54.32±0.49 &
  53.65±0.30 &
  \textcolor{blue}{\bf{52.32±0.29}} &
  53.63±0.51 &
  54.18±0.22 &
  \bf{53.51±0.32} & \\ \cline{3-12}
  &
  1.12\% &
  53.49±0.22 &
  53.68±0.27 &
  \bf{53.40±0.26} &
  53.79±0.48 &
  53.81±0.30 &
  54.45±0.30 &
  \textcolor{blue}{\bf{52.80±0.95}} &
  53.78±0.40 &
  54.00±0.51 &
  53.89±0.58 &
  \\ \bottomrule
\end{tabular}%
}
\end{table}

Although condensed graphs reconstruct graph structures by significantly changing node features and graph connections, the label information is still preserved. 
This preserved information on condensed graphs can be exploited to attack the original graph data by training detection models to infer the potential membership of nodes or links in the original networks. 
To comprehensively evaluate privacy preservation, we apply Membership Inference Attacks (MIA) and Link Membership Inference Attacks (LMIA) to assess the condensed graphs from both node and link privacy perspectives.

The experimental results for LMIA are presented in \autoref{tab:lmia}. 
It demonstrates that GDEM achieves the best performance in link privacy preservation across nearly all datasets, attributed to its eigenbasis-matching method, which avoids preserving link information through GNN-based learning.
Compared to the original data, almost all methods exhibit better privacy preservation performance in link inference tasks.
Specifically, HyDRO\textsuperscript{+} achieves the best results on the Automotive dataset at a 0.11\% reduction rate and delivers comparable performance on the VAT dataset at a 3.00\% reduction rate.
Furthermore, compared to methods like SGDD and HyDRO, which excel in link prediction tasks, HyDRO\textsuperscript{+} outperforms them in preserving link membership privacy, achieving lower F1 scores across almost all datasets with diverse reduction rates. 
This highlights the ability of HyDRO\textsuperscript{+} in balancing privacy preservation and utility, effectively mitigating LMIA risks  while maintaining strong performance in link prediction tasks.
Overall, HyDRO\textsuperscript{+} demonstrates the best trade-off between reducing LMIA risks and preserving model performance, making it a robust choice for privacy-aware graph condensation.

To further evaluate the privacy preservation of condensed graphs, we also apply Membership Inference Attacks (MIA) to assess whether the condensed graphs can be exploited to infer node labels from the original graphs. The results in \autoref{tab:lmia} show that condensed graphs generated by traditional methods and structure-based graph condensation achieve comparable or superior privacy preservation compared to the original graphs.
However, no graph condensation method consistently excels in MIA accuracy across all cases. Methods such as GDEM, DosCond, SGDD, and HyDRO\textsuperscript{+} each demonstrate the best MIA accuracy in different datasets. 
Notably, methods that perform well in link prediction tasks, such as HyDRO\textsuperscript{+} and SGDD, also exhibit strong results in the MIA test, outperforming HyDRO in most cases. 
This suggests that HyDRO\textsuperscript{+} strikes an optimal balance between utility for link prediction and privacy preservation for nodes and links, effectively mitigating the risk of node label and connection inference attacks while maintaining high performance in downstream tasks.

\subsection{Computational and Storage Efficiency}

\begin{table}[ht]
\renewcommand{\arraystretch}{1.2}
\centering
\caption{Comparison of computational efficiency and storage size between the original graphs and the condensed graphs generated by HyDRO\textsuperscript{+}.}
\label{tab:efficiency}
\resizebox{.8\textwidth}{!}{%
    \begin{tabular}{ccccccccc}
    \toprule
    \multirow{2}{*}{} &
      \multicolumn{2}{c}{\textbf{Computers ($r$=.4\%)}} &
      \multicolumn{2}{c}{\textbf{Photo ($r$=.8\%)}} &
      \multicolumn{2}{c}{\textbf{VAT ($r$=1.5\%)}} &
      \multicolumn{2}{c}{\textbf{Automotive ($r$=.11\%)}} \\  \cmidrule(lr){2-3} 
      \cmidrule(lr){4-5} 
      \cmidrule(lr){6-7} 
      \cmidrule(lr){8-9}
      \multirow{2}{*}{} &
      Original &
      \cellcolor[HTML]{EFEFEF}Condensed &
      Original &
      \cellcolor[HTML]{EFEFEF}Condensed &
      Original &
      \cellcolor[HTML]{EFEFEF}Condensed &
      Original &
      \cellcolor[HTML]{EFEFEF}Condensed \\ \hline
    Speed (s) &
      120.43 &
      \cellcolor[HTML]{EFEFEF}6.11 &
      74.88 &
      \cellcolor[HTML]{EFEFEF}7.71 &
      31.06 &
      \cellcolor[HTML]{EFEFEF}5.48 &
      73.76 &
      \cellcolor[HTML]{EFEFEF}5.89 \\
    Storage (MB) &
      78.2 &
      \cellcolor[HTML]{EFEFEF}.177 &
      41.7 &
      \cellcolor[HTML]{EFEFEF}1.2 &
      1.1 &
      \cellcolor[HTML]{EFEFEF}.0135 &
      8.9 &
      \cellcolor[HTML]{EFEFEF}.0141 \\ \hline
    \end{tabular}%
}
\end{table}

For computational efficiency, we record the link prediction model training time over 1000 epochs on four network datasets and compare the time efficiency between the original graphs and the condensed graphs generated by HyDRO\textsuperscript{+}. 
The results in \autoref{tab:efficiency} show that training on the condensed graphs generated by HyDRO\textsuperscript{+} is nearly 20 times faster than training on the original Computers network dataset.
Additionally, we demonstrate that our condensed graphs can be approximately 452.4 times smaller in storage size compared to the original Computers network dataset. Furthermore, we achieve at least 9.71 times faster training times on the VAT network dataset and 34.75 times less storage efficiency on the Photo network dataset compared to using the original data. 
These results highlight the significant computational and storage advantages of our approach across diverse network datasets.

\section{Discussion and Implications}
\label{sec:discussion}

In this study, we introduce the use of condensed graphs to replace original graphs, enabling privacy-preserved link prediction information sharing in real-world complex networks. 
Our proposed graph condensation method---HyDRO\textsuperscript{+}---leverages algebraic Jaccard similarity to enhance the initialization process of condensed graphs. 
Unlike random sampling, which ignores the underlying graph structure, algebraic Jaccard prioritizes structurally significant nodes, ensuring that sampled subgraphs better approximate the spectral properties of the original graph. 
This approach captures both local and global graph topology, making it highly effective for link prediction tasks. 
Additionally, algebraic Jaccard is robust to noise and sparsity, making it suitable for large-scale and complex networks where random sampling often fails to capture essential patterns.
HyDRO\textsuperscript{+} embeds the selected important node features from the original graph into hyperbolic space, predicting graph connectivity on the surface of the Poincaré ball. 
This approach generates a synthetic graph that more accurately preserves the topological properties of the original graph, enhancing its performance for downstream tasks such as link prediction. By overcoming the limitations of random sampling, our method improves the accuracy of synthetic graph generation and ensures that critical structural properties are retained.

In addition, HyDRO\textsuperscript{+} can also enhance privacy preservation compared to random sampling.
First, by prioritizing nodes based on their structural roles rather than selecting them randomly, it minimizes the possibility of select highly sensitive or identifiable nodes from the whole graph. 
Second, it preserves the overall structural diversity of the graph, ensuring that samples are drawn evenly from various regions. This balanced approach prevents biased subgraphs that could reveal localized sensitive patterns, thereby strengthening privacy preservation while maintaining graph integrity.
Furthermore, Algebraic Jaccard sampling is less sensitive to noise and outliers in the graph. Random sampling can inadvertently amplify noise or outliers, which may contain sensitive information. By focusing on structural similarity, Algebraic Jaccard sampling filters out irrelevant or noisy connections, resulting in a cleaner and more privacy-preserving condensed graph.

However, further experimental evaluation is necessary to validate the effectiveness of our method. 
Specifically, evaluations on larger-scale datasets comprising hundreds of millions of nodes, as well as on a broader range of complex networks from domains such as biology and finance, are essential to comprehensively assess its scalability and generalizability.
Such studies will help ensure the robustness of our approach across diverse applications and varying data scales.
This will help ensure the generalizability and robustness of our approach across diverse applications and data scales. Besdies, the performance is sensitive to hyperparameter selection, often requiring extensive tuning to adapt to different graph structures. Also, the method assumes a static graph structure, which limits its applicability to dynamic or temporal graphs without significant modifications. Finally, the interpretability of the learned hyperbolic embeddings remains an open challenge, especially in understanding how specific geometric properties influence downstream task performance. These limitations present opportunities for future research and improvement.

In real-world scenarios, the synthetic network distilled by HyDRO\textsuperscript{+} offers significant advantages in privacy-enhanced data sharing and computational efficiency. Firstly, it enables organizations to share link prediction insights derived from condensed graphs without exposing raw data, ensuring sensitive information remains confidential. For instance, companies in supply chain networks can collaboratively analyze risks or optimize logistics using synthetic graphs, avoiding the disclosure of proprietary data. Similarly, retailers can leverage synthetic co-purchasing networks to predict product dependencies and optimize inventory management without revealing detailed customer purchase histories. This privacy-preserving approach fosters secure collaboration across industries.
Secondly, by distilling essential information into a compact format, HyDRO\textsuperscript{+} significantly reduces computational and storage burdens compared to processing large-scale raw data. This efficiency allows organizations to perform complex analyses, such as link prediction, with fewer resource requirements. The method is particularly valuable for industries handling massive, dynamic datasets, as it enables efficient updates to condensed graphs without reprocessing entire networks. Together, these benefits---privacy-enhanced sharing and computational efficiency---make HyDRO\textsuperscript{+} a promising tool for industries reliant on dynamic and sensitive network data.

\section{Conclusion and Future Work}
\label{sec:conclusion}

We present HyDRO\textsuperscript{+}, an improved version of HyDRO \citep{long2025random} for graph condensation, which improves upon its predecessor by replacing random node selection with a guided selection based on the algebraic Jaccard similarity.
Extensive experiments show that HyDRO\textsuperscript{+} achieves overall better performance than state-of-the-art methods under the proposed evaluation framework.
Notably, HyDRO\textsuperscript{+} achieves the best balance between link prediction accuracy and privacy preservation while significantly reducing computational time (by an order of magnitude) and storage usage (by two orders of magnitude).
These advantages make it well-suited for real-world applications in large-scale networks.

This work provides insights into the domain of complex networks, where data sharing is necessary for graph-based operations such as link prediction and privacy protection must be guaranteed.
By generating condensed graphs, HyDRO\textsuperscript{+} facilitates collaboration between organizations, enabling researchers and practitioners to gain collective insights without compromising sensitive information. 
Besides, this capability is particularly valuable in privacy-sensitive domains, such as supply chains, finance, and healthcare, where achieving a balance between data privacy and utility is critical.

Our future work will focus on evaluating HyDRO\textsuperscript{+} on larger and more diverse datasets. 
Specifically, we will conduct experiments on networks with hundreds of millions of nodes are necessary to validate the effectiveness and scalability of our approach in real-world scenarios. 
Additionally, we will extend our evaluation to other domains, such as biology and finance, assessing the generalizability and robustness of our approach across a broader range of applications. 
These efforts will further validate whether HyDRO\textsuperscript{+} is a versatile and scalable solution for privacy-preserving graph-based downstream tasks, paving the way for safer and more efficient use of network data in sensitive industrial domains.

\newpage
\appendix

{\small
\bibliographystyle{plainnat}
\bibliography{reference}
}
\end{document}